\documentclass[aos]{imsart}

\RequirePackage{amsthm,amsmath,amsfonts,amssymb}
\RequirePackage[numbers]{natbib}
\RequirePackage[colorlinks,citecolor=blue,urlcolor=blue]{hyperref}
\RequirePackage{graphicx}
\usepackage{amsmath}
\usepackage{tikz}
\usetikzlibrary{automata,arrows,positioning,calc}
\usepackage{verbatim}

\usepackage[linesnumbered,ruled,vlined]{algorithm2e}
\SetKwInput{KwInput}{Input}                
\SetKwInput{KwOutput}{Output}              

\startlocaldefs
\theoremstyle{plain}

\newtheorem{theorem}{Theorem}[section]
\newtheorem{lemma}[theorem]{Lemma}

\newtheorem{remark}[theorem]{Remark}

\theoremstyle{remark}
\newtheorem{definition}[theorem]{Definition}

\endlocaldefs

\begin{document}

\begin{frontmatter}
\title{Markov Observation Models}
\runtitle{Markov Observation Models}

\begin{aug}
\author[A]{\fnms{Michael}~\snm{ A. Kouritzin}\ead[label=e1]{michaelk@ualberta.ca}},
\address[A]{Department of Mathematical and Statistical Sciences,
University of Alberta\printead[presep={,\ }]{e1}}

\end{aug}

\begin{abstract}
Herein, the Hidden Markov Model is expanded to allow for Markov
chain observations.
In particular, the observations are assumed to be a Markov chain
whose one step transition probabilities depend upon the hidden
Markov chain.
An Expectation-Maximization analog to the Baum-Welch algorithm is
developed for this more general model to estimate the transition
probabilities for both the hidden state and for the observations
as well as to estimate the probabilities for the initial joint 
hidden-state-observation distribution.
A believe state or filter recursion to track the hidden state
then arises from the calculations of this Expectation-Maximization algorithm.
A dynamic-programming analog to the Viterbi algorithm is also
developed to estimate the most likely sequence of hidden states
given the sequence of observations.
\end{abstract}



\end{frontmatter}

\setcounter{equation}{0}

\section{Introduction}\label{Intro}

Hidden Markov models (HMMs) were introduced in a series of papers by Baum and 
collaborators \cite{Baum1}, \cite{Baum2}.
Traditional HMMs have enjoyed termendous success in 
applications like computational finance \cite{Petro}, 
single-molecule kinetic 
analysis \cite{Nico}, animal tracking \cite{Sidrow}, forcasting commodity futures \cite{Rogemar} and protein folding \cite{Stig}.
%
In HMMs the unobservable hidden states {$\displaystyle X $} are a discrete{-}time Markov chain and the observations process {$\displaystyle Y $} is some distorted, corrupted partial information or measurement of the current state of {$\displaystyle X $} satisfying the condition
\begin{center}{$\displaystyle P\left(Y_{n}\in A\big|X_{n},X_{n-1},...,X_1\right)=P\left(Y_{n}\in A\big|X_{n}\right). $}\end{center}
These probabilities, {$\displaystyle P\left(Y_{n}\in A\big|X_{n}\right) $}, are called the {\em emission probabilities}.

This type of observation modeling can be limiting.  
Consider observations $Y$ consisting of daily stock price and volume 
that are based upon a hidden (bullish/bearish type) market state $X$.  
If there was really just an emission probability, the prior day's price
and volume would be completely forgotten and a new one would be
chosen randomly only depending solely upon the market bull/bear state.
Clearly, this is not what happens.
The next day's price and volume is related to the prior day's in some way.
Perhaps, prices are held in a range by recent earnings or volume is 
elevated for several days due to some company news.
Indeed, the autoregressive HMM (AR-HMM) was been introduced because
the (original) HMM does not allow for an observation to depend upon
a past observation.
For the AR-HMM the observations take the structure:
\begin{equation}\label{ARHMM}
Y_n=\beta_0^{(X_n)}+\beta_1^{(X_n)}Y_{n-1}+\cdots+\beta_p^{(X_n)}Y_{n-p}
+\varepsilon_n,
\end{equation}
where $\{\varepsilon_n\}_{n=1}^\infty$ are a (usually zero-mean Gaussian)
i.i.d.\ sequence of random variables and the autoregressive coefficients
are functions of the current hidden state $X_n$.
Most critically, one might view this AR-HMM as a linear, Gaussian partial patch to
the HMM deficiency and expect a more general, useful theory.
Still, the AR-HMM has experienced strong success in applications
like speech recognition
(see \cite{BrLe}), diagnosing blood infections (see \cite{Stanc})
and the study of climate patterns (see \cite{Xuan}).
Finally, there are general models that truly incorporate (possibly
non-linear) dependencies of $Y_n$ on past values of $Y$ referred
to as {\em Markov-switching models} or sometimes {\em Markov jump systems}.
These are very general models that are particularly important in
financial applications.
However, as mentioned in \cite{CaMoRy} the analyses of Markov-switching
models can be far more intricate than those of HMM due to the fact that the 
properties of the observed
process are not directly controlled by those of the hidden chain.

It is perhaps easiest to explain our work in the context of the most
general \emph{Pairwise Markov Chain} (PMC) model from \cite{Piec}.
In \cite{Piec}, it was only assumed that $(X,Y)$ was jointly Markov and 
important formula for Bayesian Maximal Posterior Mode restoration 
were still derived.
However, when it came to parameter estimation from incomplete data
it was realized that the Baum-Welch algorithm could not be generalized to 
this most general PMC setting and, instead, the general Iterative
Conditional Estimation was resorted to.
Likewise no Viterbi-like algorithm exists for finding the most likely
sequence from an observed data sequence of a PMC.
Our goal is to narrow the gap between the limited HMM and AR-HMM where
Baum-Welch and Viterbi algorithms are known and the practically-important
PMC which has no such algorithms by introducing a model that falls between
the two that still has these algorithms.  
In particular, 
we will establish Baum-Welch-like and Viterbi-like algorithms
for estimating (initial and transition) probabilities and the most likely
sequence from the observed data for a new, still practially
important model in the discrete setting.
We refer to our models as Markov Observation Models (MOM).

Perhaps, the most important goals of HMM are calibrating the model, real-time 
believe state propagation, i.e. filtering, and decoding the whole hidden
sequence from the observation sequence.  
The first problem is solved mathematically in the HMM setting by the 
Baum{-}Welch re-estimation algorithm, which is an application of the 
Expectation{-}Maximization (EM) algorithm, predating the EM algorithm.  
The filtering problem is also solved effectively using a recursive algorithm
that is similar to part of the Baum-Welch algorithm.  
In practice, there can be numeric problems like a multitude of local maxima to trap the Baum{-}Welch algorithm or inefficient matrix operations when the state size is large but the hidden state resides in a small subset most of the time.  In these cases, it can be adviseable to use particle filters or other alternative methods, which are not the subject of this note (see instead \cite{CaMoRy} for
more information).  
The forward and backward propagation probabilities of the Baum-Welch algorithm
also tend to get very small over time.
While satisfactory results can sometimes be obtained by (often logarithmic)
rescaling, this is still a severe problem limiting the
use of the Baum-Welch algorithm (see more explanation within).
Our raw algorithms for the more general Markov observation models will also 
share these difficulties but as a secondary
contribution we will explain how to avoid this \emph{small number problem}
when we give our final pseudocode so our EM algorithm will truly apply
to many \emph{big data} problems.

The optimal complete-observation sequence decoding problem in the HMM case is solved by the Viterbi
algorithm (see \cite{Viterbi}, \cite{Rabiner}), which is a dynamic programming type algorithm.
Given the sequence of observations $\{Y_i\}_{i=1}^N$ and the model probabilities, the
Viterbi algorithm returns the most likely hidden state sequence
$\{X^*_i\}_{i=1}^N$.
The Viterbi algorithm is a forward-backward algorithm like the Baum-Welch algorithm
and hence computer efficient but not real time. 
The most natural applications of the Viterbi algorithm are perhaps
speech recognition \cite{Rabiner} and text recognition \cite{Shinghai}.
We develop a Markov observation model generalization to the Viterbi algorithm
and explain how to handle the small number problem in this algorithm as well.

The HMM can be thought of as a nonlinear generalization of the
earlier Kalman filter (see \cite{Kalman}, \cite{KalBucy}).
Nonlinear filtering theory is another related generalization of
the Kalman filter and has many cellebrated successes like
the Fujisaki-Kallianpur-Kunita and the Duncan-Mortensen-Zakai equations
(see e.g. \cite{Zakai}, \cite{FKK72}, \cite{KuOc88} for some of the original 
work and \cite{KoLo08}, \cite{KuNa} for some of the more recent general 
results).
The hidden state, called signal in nonlinear filtering theory, 
can be a general Markov process model and live in a general state
space but there is no universal EM algorithm for identifying
the model like the Baum-Welch algorithm nor dynamic programming 
algorithm for identifying a most likely hidden state path like the
Viterbi algorithm.
Rather the goals are usually to compute filters, predictors and smoothers,
for which there are no exact closed form solutions, except in isolated
cases (see \cite{Ko98}), and approximations have to be used.
Like HMM, nonlinear filtering has enjoyed widespread application.
For instance, the subfield of nonlinear particle filtering, also known
as sequential Monte Carlo, has a number of powerful algorithms
(see \cite{Pitt}, \cite{DMKoMi}, \cite{Ko17a}, \cite{Chop20}) and has been applied to numerous problems in areas like
bioinformatics \cite{Haji}, economics and mathematical finance \cite{Creal}, 
intracellular movement \cite{MaNe}, fault detection \cite{DAmato},
pharmacokinetics \cite{Bonate} and many other fields.
Still, like HMM, the observations in nonlinear filter models are largely
limited to distorted, corrupted, partial observations of the signal
with very few limited exceptions like \cite{CrKoXi}.  

The purpose of this note is to promote a class of Markov Observation Models (MOM) that will be shown to subsume the HMM and AR-HMM models
in the next section.  
MOM is also very different than the models considered in non-linear filtering.
Hence, to the author's knowledge, MOM represents a practically important
class of models to analyze and apply to real world problems.
Both the Baum-Welch and the Viterbi algorithms will be extended to
these MOM models as, together with the model itself, the main contributions.
A real-time filtering recursion is also extended.
It should be noted that our EM and dynamic programming generalizations
of the Baum-Welch and Viterbi algorithms include new methods for handling
an unseen first observation that is not even part of the HMM model.
Finally, the small number problem encountered in HMM and the raw MOM algorithms
is resolved.

The layout of this note is as follows.
In the next section, we give our model as well as our main notation.  
In Section \ref{ProbEM}, we apply EM techniques to derive an analog to the Baum{-}Welch algorithm for identifying the system (probability) parameters.  
In particular, joint recursive formulas for the hidden state transition
probabilities, observation transition probabilities and the initial
joint hidden-observation state distribution are derived.
Section \ref{EMSmall} translates these formula into a pseudocode 
implementation of our EM algorithm.
More calculations and explanations are included to explain how we
avoid the small number problem often encountered in HMM.
Section \ref{ConProb} is devoted to connecting the limit points of the EM type 
algorithm to the maxima of the conditional likelihood given the observations.
Section \ref{Viterbi} contains our real-time filter process 
recursion and our forward-backward most likely hidden sequence detection.  
Specifically, it contains our dynamic programming analog to the
Viterbi algorithm for MOM as well as its derivation and pseudocode
implementation.
Finally, Section \ref{BitCoin} features a application of our (Baum-Welch-like) EM and
our (Viterbi-like) dynamic programming algorithms on real
bitcoin data to detect uptrends.

\section{Model}\label{Model}

Suppose $N$ is some positive integer (representing the final time) and {$\displaystyle O $} is some discrete observation space. 
In our model, like HMM, the hidden state is a homogeneous Markov chain 
{$\displaystyle X $} on some discrete (finite or countable)
state space {$\displaystyle E $} with one step transition probabilities denoted by 
{$\displaystyle p_{x\rightarrow {x^{\prime}}} $} for {$\displaystyle x,{x^{\prime}}\in E $}.  
However, in contrast to HMM, we allow self dependence in the observations.  
(This is illustrated by right arrows between the 
$Y$'s in Figure \ref{fig C3} below.)
In particular, given the hidden state $\{X_i\}_{i=0}^N$, we take the 
observations to be a (conditional) Markov chain $Y$ with transitions 
probabilities
\begin{equation}
\displaystyle P\!\left(Y_{n+1}=y\Big|\{X_{i}=x_{i}\}_{i=0}^{n+1},\{Y_{j}=y_{j}\}_{j=0}^{n}\right)=q_{y_{n}\rightarrow y}(x_{n+1})\ \forall x_0,...,x_N\in E;\ y,y_n\in O\!
\end{equation}
that do not affect the hidden state transitions in the sense
\begin{equation}\label{Xdependsim}
P(X_{n+1}=x'\Big| X_n=x,\{X_i\}_{i<n}, \{Y_j\}_{j\le n})=p_{x\rightarrow {x^{\prime}}},\ \forall x,x^{\prime}\in E,n\in \mathbb N_0
\end{equation}
still.  
This means that 
\begin{equation}\label{Ydependsim}
P\left(Y_{n+1}=y\Big|\{X_{i}\}_{i=0}^{n+1},\{Y_j\}_{j\le n}\right)=P\left(Y_{n+1}=y\Big|X_{n+1},Y_{n}\right),\ \forall y\in O
\end{equation}
i.e. that the new observation only depends upon the new hidden state
(as well as the past observation),
and also that the hidden state, observation pair {$\displaystyle \left(\begin{array}{c}X\\Y\end{array}\right) $} is jointly Markov (in addition to the hidden state itself being Markov) with joint one step transition probabilities
\begin{center}{$\displaystyle P\left(X_{n+1}=x,Y_{n+1}=y\Big|X_{n}=x_{n},Y_{n}=y_{n}\right)=p_{x_{n}\rightarrow x}\,q_{y_{n}\rightarrow y}(x)\ \forall x,x_n\in E;\ y,y_n\in O. $}\end{center}

\begin{figure}
\tikzstyle{int}=[draw, fill=blue!20, minimum size=2em]
\begin{tikzpicture}[->, >=stealth', auto, thick]
\tikzstyle{every state}=[fill=white, draw=black, thick, text=black]
        \node[state,int]  (a) {$X_0$};
        \node[state,int]  (b) [right = 1.5cm of a]   {$X_1$};
        \node[state,int]  (c) [right = 1.5cm of b]   {$X_2$};
        \node[state,int]  (d) [right = 1.5cm of c]   {$X_3$};
        \node[state,int]  (e) [right = 1.5cm of d]   {$X_N$};
    \node (o) [left of=a,node distance=2.5cm, coordinate] {a};
       
        \node[state,int]  (f) [below = 1.25cm of a] {$Y_0$};
        \node[state]  (g) [below = 1.25cm of b] {$Y_1$};
        \node[state]  (h) [below = 1.25cm of c] {$Y_2$};
        \node[state]  (i) [below = 1.25cm of d] {$Y_3$};
        \node[state]  (j) [below = 1.15cm of e] {$Y_N$};
    \node (p) [left of=f,node distance=2.5cm, coordinate] {a};
      
        \node  (k) [below = 1cm of g] {Obs 1};
        \node  (l) [below = 1cm of h] {Obs 2};
        \node  (m) [below = 1cm of i] {Obs 3};
        \node  (n) [below = 1cm of j] {Obs N};

\path (a) edge[right] (b) (b) edge[right] (c) (c) edge[right] (d) 
      (f) edge[right] (g) (g) edge[right] (h) (h) edge[right] (i) 
      (a) edge[below] (f)
      (b) edge[below] (g)
      (c) edge[below] (h)
      (d) edge[below] (i)
      (e) edge[below] (j)
      (g) edge[below] (k)
      (h) edge[below] (l)
      (i) edge[below] (m)
      (j) edge[below] (n);
    \path[->] (o) edge node {prior $X$} (a);
    \path[->] (p) edge node {prior $Y$} (f);
\draw [dashed] (d) edge[below] (e);
\draw [dashed] (i) edge[below] (j);
\end{tikzpicture}
\begin{tabular}{r@{: }l r@{: }l}
shaded values & not observed; &$X_0,Y_0$& not part of normal HMM\\
 unshaded & observed; & $X_0, X_1, Y_0$& Estimated together in 
Viterbi 
\end{tabular}
\caption{Markov Observation Model Structure}
\label{fig C3}
\end{figure}
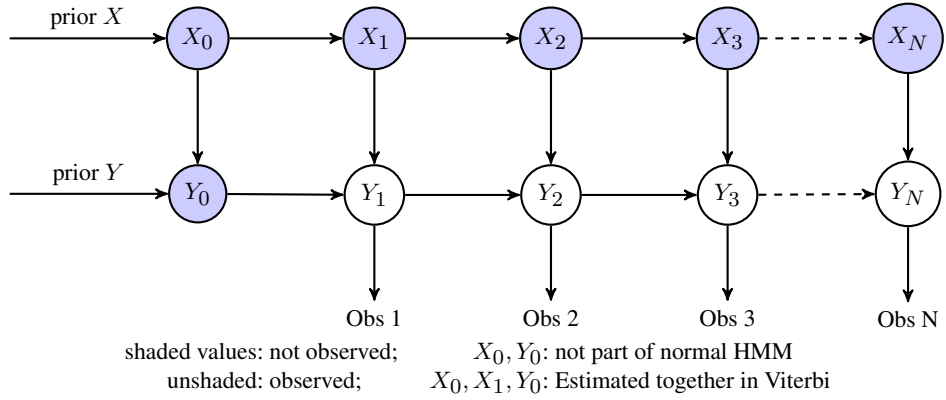

The joint Markov property then implies that 
\begin{center}{$\displaystyle P\left(X_{n+1}=x,Y_{n+1}=y\Big|X_{1}=x_{1},Y_{1}=y_{1},X_{2}=x_{2},Y_{2}=y_{2},...,X_{n}=x_{n},Y_{n}=y_{n}\right)=p_{x_{n}\rightarrow x}q_{y_{n}\rightarrow y}\left(x\right). $}\end{center}
Notice that this generalizes the emisson probability to
\begin{center}{$\displaystyle P\left(Y_{n}\in A\big|X_{n},X_{n-1},...,X_1;Y_{n-1},...,Y_1\right)=P\left(Y_{n}\in A\big|Y_{n-1},X_{n}\right)=\sum_{y\in A}q_{Y_{n-1}\rightarrow y}\left(X_n\right) $}
\end{center}
so MOM generalizes HMM by just taking $q_{Y_{n-1}\rightarrow y}\left(X_n\right)=b_{X_{n}}(y)$, a state dependent probability mass function.
To see that MOM generalizes AR-HMM, we re-write (\ref{ARHMM}) as
\begin{equation}
\underbrace{\left[
\begin{array}{c} Y_n\\Y_{n-1}\\Y_{n-2}\\\vdots\\Y_{n-p+1}\end{array}
\right]}_{\mathcal Y_n}
=
\left[
\begin{array}{ccccc}\beta_1^{(X_n)}&\beta_2^{(X_n)}&\beta_3^{(X_n)}&\cdots&\beta_p^{(X_n)}\\
1&0&0&\cdots&0\\
0&1&0&\cdots&0\\
\vdots&&\ddots&\vdots\\
0&0&0&\cdots 1&0
\end{array}
\right]
\underbrace{\left[
\begin{array}{c} Y_{n-1}\\Y_{n-2}\\Y_{n-3}\\\vdots\\Y_{n-p}\end{array}
\right]}_{\mathcal Y_{n-1}}
+\left[
\begin{array}{c} \beta_0^{(X_n)}+\varepsilon_n\\0\\0\\\vdots\\0\end{array}
\right],
\end{equation}
which, given the hidden state $X_n$, gives an explicit formula for 
$\mathcal Y_{n}$ in terms of only $\mathcal Y_{n-1}$ and some
independent noise $\varepsilon_n$.
Hence, $\{\mathcal Y_{n}\}$ is obviously conditionally Markov
and $\{(X_n,\mathcal Y_{n})\}$ is a MOM.

A subtly that arises with our Markov Observation Model (MOM) over HMM is 
that we need an enlarged initial distribution since we have a $Y_0$ that
is not observed (see Figure \ref{fig C3}).
Rather, we think of starting up the observation process at time $1$
even though there were observations to be had prior to this time.
Further, since we generally do not know the 
model parameters, we need means to estimate this initial distribution
\begin{center}{$\displaystyle P\left(X_{0}\in dx_{0},Y_{0}\in dy_{0}\right)=\mu \left(dx_{0},dy_{0}\right) $}.\end{center}

It is worth noting that our model resembles the stationary PMC under 
Condition (H) in \cite{Piec}, which forces the Hidden state to be Markov
by Proposition 2.2 of \cite{Piec}.

%
%
\subsection{Key Notation}

\begin{itemize}
\item
We will use the shorthand notation {$\displaystyle P\left(Y_{1},...,Y_{n}\right) $} for {$\displaystyle P\left(Y_{1}=y_1,...,Y_{n}=y_n\right)|_{y_1=Y_1,...,y_n=Y_n} $}.

\item $\alpha^k_n(x)=P^k(X_n=x,Y_{1},...,Y_{n})$ and 
$\beta^k_n(x)=P^k(Y_{n+1},...,Y_{N}\big|X_n=x,Y_{n})$ 
(both defined differently when $n=0$ below) are probabilities
computed using the current estimates $p^k_{x\rightarrow x'}$, 
$q^k_{y\rightarrow y'}(x)$ and $\mu^k(x,y)$ of the transition
and initial probabilities. 
$\alpha^k_n(x)$ and $\beta^k_n(x)$ will be key variables in the forward 
respectively backward propagation step of our \emph{raw} Baum-Welch-like
EM algorithm for estimating the transition and initial probabilities.
For notational ease, we will drop the fact $P$ depends on $k$ hereafter.

\item The filter $\pi^k_n(x)=P(X_n=x\big|Y_{1},...,Y_{n})$ and
$\chi^k_n(x)=\beta^k_n(x)\frac{P(Y_{1},...,Y_{n})}{P(Y_{1},...,Y_{N-1})}$
are used in our \emph{refined} Baum-Welch-like algorithm to replace $\alpha_n^k$
respectively $\beta^k_n$ of the raw algorithm in order to solve the 
small number problem discussed below.
Whereas $\alpha^k_n(x)\beta^k_n(\xi)$ is often the product of two tiny
unequally sized factors, $\pi^k_n$ and $\chi^k_n$ are scaled to always
be manageable factors.
Yet, $\pi^k_n(x)\chi^k_n(\xi)=\frac{\alpha^k_n(x)\beta^k_n(\xi)}{P(Y_{1},...,Y_{N-1})}$
and both $\pi^k_n$ and $\chi^k_n$ satisfy nice forward and backward 
recursions so they are efficient to compute and our refined EM algorithm
for MOM is efficient and avoids the small number problem.

\item $\delta_n(x)=\max\limits_{y_0;x_0,x_1,...,x_{n-1}}\!\!P(Y_0=y_0;X_0=x_0,X_1=x_1,...,X_{n-1}=x_{n-1};X_n=x;Y_1,...,Y_n)$ is the key internal function in our
Viterbi-like dynamic programming algorithm for determining the most
likely sequence of hidden states.
$\delta_n$ also suffers from the small number problem as it tends to 
get ridiculously small as $n$ increases.
However, since there is only one factor it is easy to scale and
scaling each $\delta_n$ does not affect Viterbi-like algorithm,
we can replace $\delta_n$ with a properly scaled version $\gamma_n$
below.

%
\end{itemize}

\section{Probability Estimation via EM algorithm}\label{ProbEM}

In this section, we develop a recursive expectation-maximum algorithm
that can be used to create convergent estimates for the transition
and initial probabilities of our MOM models.
We leave the theoretical justification of convergence to Section \ref{ConProb}.

The main goal of developing an EM algorithm would be to find {$\displaystyle p_{x\rightarrow x'} $} for all {$\displaystyle x,x'\in E $}, {$\displaystyle q_{y\rightarrow y'}(x) $} for all {$\displaystyle y,y'\in O $}, $x\in E$ and 
$\mu(x,y)$ for all {$\displaystyle x\in E $}, $y\in O$.  
Noting every time step is considered to be a transition in a discrete-time
Markov chain, we would ideally set:

\begin{eqnarray}
\displaystyle 
p_{x\rightarrow {x^{\prime}}}&=&\frac{\mbox{transitions } x\mbox{ to }{x^{\prime}}}{\mbox{ occurrences of }x} 
\\
\displaystyle q_{y\rightarrow y'}(x)&=&\frac{\mbox{transitions } y\mbox{ to }{y'}\mbox{ when } x\mbox{ is true}}{\mbox{ occurrences of }y\mbox{ when } x\mbox{ is true}} .
\end{eqnarray}
Here, `when {$\displaystyle x $} is true' means when the hidden state is in state {$\displaystyle x $}.  
However, we can never see {$\displaystyle x $} nor {$\displaystyle {x^{\prime}} $} in MOM from our data so we must estimate when they are true.  
Hence, we replace the above with

\begin{eqnarray}\label{xexpected}
\ \ \displaystyle p_{x\rightarrow {x^{\prime}}}&\!=&\!\frac{\mbox{Expected transitions } x\mbox{ to }{x^{\prime}}}{\mbox{Expected occurrences of }x}
=\frac{\sum\limits_{n=1}^{N}P(X_{n-1}=x,X_{n}=x'\Big|Y_1,...,Y_N)}{\sum\limits_{n=1}^{N}P(X_{n-1}=x\Big|Y_1,...,Y_N)}
\end{eqnarray}
\begin{eqnarray}
&& \displaystyle q_{y\rightarrow y'}(x)\label{yexpected}=\frac{\mbox{Expected transitions } y\mbox{ to }{y'}\mbox{ when } x\mbox{ is true}}{\mbox{Expected occurrences of }y\mbox{ when } x\mbox{ is true}}\\\nonumber
&\!=&\!\frac{1_{Y_{1}=y'}P(Y_0=y,X_{1}=x\Big|Y_1,...,Y_N)+
\sum\limits_{n=2}^{ N}1_{Y_{n-1}=y,Y_{n}=y'}P(X_{n}=x\Big|Y_1,...,Y_N)}
{P(Y_0=y,X_{1}=x\Big|Y_1,...,Y_N)+\sum\limits_{n=2}^{ N}1_{Y_{n-1}=y}P(X_{n}=x\Big|Y_1,...,Y_N)}, 
\end{eqnarray}
which means we must compute {$\displaystyle  P(Y_0=y,X_{1}=x\Big|Y_1,...,Y_N)$}, {$\displaystyle P(X_{n}=x\Big|Y_1,...,Y_N) $} for all $0\le n\le N$ and {$\displaystyle P(X_{n-1}=x,X_{n}=x'\Big|Y_1,...,Y_N) $} for all {$\displaystyle 1\le n\le N $} to get these two transition probability estimates.
However, let 
\begin{equation}\label{alphadef}
\left\{
\begin{array}{lll}
\displaystyle 
\alpha _{0}\left(x,y\right)&=&P\left(Y_{0}=y,X_{0}=x\right) \\
\alpha _{n}\left(x\right)&=&P\left(Y_{1},...,Y_{n},X_{n}=x\right),
\ 1\le n\le N 
\end{array}
\right. 
\end{equation}
and
\begin{equation}\label{betadef}
\left\{
\begin{array}{lll}
\displaystyle \beta _{0}\left(x_{1},y\right)&=&P\left(Y_{1},...,Y_{N}\Big|X_{1}=x_{1},Y_{0}=y\right) \\
\displaystyle \beta _{n}\left(x_{n+1}\right)&=&P\left(Y_{n+1},...,Y_{N}\Big|X_{n+1}=x_{n+1},Y_{n}\right),\ \forall 0<n<N-1\\
\displaystyle \beta _{N-1}\left(x_N\right)&=&P\left(Y_{N}\Big|X_{N}=x_{N},Y_{N-1}\right)=q_{Y_{N-1}\rightarrow Y_N}(x_N)
\end{array}
\right. .
\end{equation}
Notice we include an extra variable $y$ in $\alpha_0,\beta_0$.
This is because we do not see the first observation $Y_0$ so we have
to consider all possibilities and treat it like another hidden state.
Now, by Bayes' rule, (\ref{betadef}) and (\ref{alphadef})
\begin{eqnarray}\label{Y0X1}
&&P(Y_0=y,X_{1}=x\Big|Y_1,...,Y_N)\\\nonumber
 & = & \displaystyle \frac{P(Y_1,...,Y_N\big| X_1=x,Y_0=y)P(X_1=x,Y_0=y)}{P(Y_1,...,Y_N)} \\\nonumber
 & = & \displaystyle \frac{\beta_{0}(x,y)\sum_{x_0}p_{x_0\rightarrow x}\alpha_0(x_0,y)}{\sum_{\xi}\alpha_N(\xi)}  .
\end{eqnarray}
Next, by the Markov property and (\ref{alphadef})
\begin{eqnarray}\label{XnXnm1p}
&&P(X_{n-1}=x,X_{n}=x'\Big|Y_1,...,Y_N)\\\nonumber
 & = & \displaystyle \frac{P(X_{n-1}=x,X_{n}=x',Y_1,...,Y_N)}{P(Y_1,...,Y_N)} \\\nonumber
 & = & \displaystyle \frac{\alpha_{n-1}(x)P(X_{n}=x',Y_{n},...,Y_N\Big|X_{n-1}=x,Y_1,...,Y_{n-1})}{P(Y_1,...,Y_N)}  \\\nonumber
 & = & \displaystyle \frac{\alpha_{n-1}(x)P(X_{n}=x',Y_{n},...,Y_N\Big|X_{n-1}=x,Y_{n-1})}{P(Y_1,...,Y_N)}  
\end{eqnarray}
so by (\ref{Xdependsim},\ref{Ydependsim},\ref{betadef},\ref{alphadef})
\begin{eqnarray}\label{XnXnm1}
&&P(X_{n-1}=x,X_{n}=x'\Big|Y_1,...,Y_N)\\\nonumber
 & = & \displaystyle \frac{\alpha_{n-1}(x)P(X_{n}=x',Y_{n},...,Y_N,X_{n-1}=x,Y_{n-1})P(X_{n}=x',X_{n-1}=x,Y_{n-1})}{P(Y_1,...,Y_N)P(X_{n}=x',X_{n-1}=x,Y_{n-1})P(X_{n-1}=x,Y_{n-1})} \\\nonumber
 & = & \displaystyle \frac{\alpha_{n-1}(x)P(Y_{n},...,Y_N\big|X_{n}=x',X_{n-1}=x,Y_{n-1})P(X_{n}=x'\big|X_{n-1}=x,Y_{n-1})}{P(Y_1,...,Y_N)} \\\nonumber
 & = & \displaystyle \frac{\alpha_{n-1}(x)P(Y_{n},...,Y_N\big|X_{n}=x',Y_{n-1})P(X_{n}=x'\big|X_{n-1}=x)}{P(Y_1,...,Y_N)} \\\nonumber
 & = &\displaystyle \frac{\alpha_{n-1}(x)\beta_{n-1}(x')p_{x\rightarrow x'}}{\sum_{\xi}\alpha_N(\xi)} 
\end{eqnarray}
for {$\displaystyle n=2,3,...,N $}.
\begin{remark}
The Baum-Welch algorithm for regular HMM also constructs the joint
conditional probability in (\ref{XnXnm1}). 
In the HMM case, the numerator in (\ref{XnXnm1}) looks like
\begin{eqnarray*}
P(X_{n-1}=x,X_{n}=x',Y_1,...,Y_N)&\!=&\!\overbrace{P(X_{n-1}=x,Y_1,...,Y_{n-1})}^{\alpha_{n-1}}
P(X_{n}=x'|X_{n-1}=x)\\
&\!*&\!\underbrace{P(Y_{n+1},...,Y_N|X_{n}=x')}_{\mbox{their }\beta_{n-1}}P(Y_n|X_{n}=x'),
\end{eqnarray*}
which works well when the observations are conditionally independent.
However, this multiplication rule does not apply in our more general
Markov observations case.
Moreover, there is no conditional independence so
\begin{eqnarray*}
&&P(X_n=x',Y_{n},...,Y_N|X_{n-1}=x,Y_{n-1})\\
&&\ne
P(X_n=x'|X_{n-1}=x,Y_{n-1})P(Y_{n},...,Y_N|X_{n-1}=x,Y_{n-1}).
\end{eqnarray*}
Our new strategy is to define
\[
\beta_{n-1}(x')=P(Y_{n},...,Y_N|X_n=x',Y_{n-1})
\]
and note that
\[
\beta_{n-1}(x')=P(Y_{n},...,Y_N|X_n=x',X_{n-1}=x,Y_{n-1})
\]
for all $x$.
Surprisingly, with such modest changes, the algorithms 
making HMM such a powerful tool translate to the more general MOM
models.
\end{remark}
It follows from (\ref{XnXnm1}) that 
\begin{equation}\label{Xn}
P(X_{n}=x\Big|Y_1,...,Y_N) =\displaystyle\alpha_{n}(x)\sum_{x_{n+1}} \frac{\beta_{n}(x_{n+1})p_{x\rightarrow x_{n+1}}}{\sum_{\xi}\alpha_N(\xi)} 
\end{equation}
for {$\displaystyle n=1,2,...,N-1$} and
\begin{equation}\label{Xn2}
P(X_{n}=x\Big|Y_1,...,Y_N)=\displaystyle \beta _{n-1}\left(x\right)\sum_{x_{n-1}}\frac{p_{x_{n-1}\rightarrow x}\alpha_{n-1}(x_{n-1})}{\sum\limits_{\xi}\alpha_N(\xi)}
\end{equation}
for {$\displaystyle n=2,3,...,N$}.
Similarly to (\ref{XnXnm1p},\ref{XnXnm1}), one has that
\begin{eqnarray} \label{X0X1}
&\!\!&\!\! P(X_{0}=x,X_{1}=x'\Big|Y_1,...,Y_N) \displaystyle \\\nonumber
&\!\!=&\!
\frac{\sum_y P(X_{0}=x,Y_0=y)P(X_{1}=x';Y_1,...,Y_N\big|X_{0}=x,Y_0=y)}{P(Y_1,...,Y_N)} \\\nonumber
&\!\!=&\! \displaystyle \frac{\sum_y\alpha_{0}(x,y)p_{x\rightarrow x'}\beta_{0}(x',y)}{\sum_{\xi}\alpha_N(\xi)}
\end{eqnarray}
and so
\begin{equation} \label{X0}
P(X_{0}=x\Big|Y_1,...,Y_N) =\displaystyle \sum_{x'}\frac{\sum_y\alpha_{0}(x,y)p_{x\rightarrow x'}\beta_{0}(x',y)}{\sum\limits_{\xi}\alpha_N(\xi)}.
\end{equation}
$\alpha_n$ and $\beta_n$ 
are computed recursively below 
using the prior estimates of {$\displaystyle p_{x\rightarrow {x^{\prime}}} $}, 
{$\displaystyle q_{y\rightarrow y^{\prime}}\left(x\right) $} and $\mu$.

Recalling that there are prior observations that we do not 
see, we must also estimate an initial joint distribution for
an initial hidden state and observation.
An expectation-maximization argument for the initial distribution
leads one to the assignment
\begin{eqnarray}\label{BayesRule}
\displaystyle\mu(x,y)&\!=&P(X_0=x,Y_0=y\Big|Y_1,...,Y_N)\\\nonumber
&\!=&\frac{P(Y_1,...,Y_N\Big|X_0=x,Y_0=y)P(X_0=x,Y_0=y)}{P(Y_1,...,Y_N)}
\end{eqnarray}
for all $x\in E$, $y\in O$, which is Bayes' rule.

Expectation-maximization algorithms use these types of formula and
prior estimates to produce better estimates.
We take estimates for {$\displaystyle p_{x\rightarrow {x^{\prime}}} $}, 
{$\displaystyle q_{y\rightarrow y'}\left(x\right) $} and $\mu(x,y)$ 
and get new estimates for these quantities iteratively using (\ref{xexpected}),
(\ref{X0X1}), (\ref{XnXnm1}), (\ref{X0}) and (\ref{Xn}):
\begin{equation}\label{empxx}
p^{\prime}_{x\rightarrow {x^{\prime}}} =
\frac{\sum\limits _y\alpha_0(x,y)p_{x\rightarrow x'}\beta_0(x',y)+\sum\limits _{n=1}^{N-1}\alpha_n(x)p_{x\rightarrow x'}\beta_n(x')}
{\sum\limits _y\sum\limits_{x_{1}}\alpha_0(x,y)p_{x\rightarrow x_{1}}\beta_0(x_1,y)+\sum\limits _{n=1}^{N-1}\sum\limits_{x_{n+1}}p_{x\rightarrow x_{n+1}}\beta _{n}\left(x_{n+1}\right)\alpha_n(x)},  
\end{equation}
then using (\ref{yexpected}), (\ref{Y0X1},\ref{Xn2})
\begin{equation}\label{emqyy}
q^{\prime}_{y\rightarrow y'}\left(x\right) =  \displaystyle \frac{1_{Y_{1}=y'}\beta _{0}(x,y)\sum\limits_{\xi}p_{\xi\rightarrow x}\alpha_{0}\left(\xi,y\right)+\sum\limits _{n=1}^{N-1}1_{Y_{n}=y,Y_{n+1}=y'}\beta _{n}\left(x\right)\sum\limits_{\xi}\alpha_{n}(\xi)p_{\xi\rightarrow x}}{\beta _{0}(x,y)\sum\limits_{\xi}p_{\xi\rightarrow x}\alpha_{0}\left(\xi,y\right)+\sum\limits _{n=1}^{N-1}1_{Y_{n}=y}\beta _{n}\left(x\right)\sum\limits_{\xi}\alpha_{n}(\xi)p_{\xi\rightarrow x}},  
\end{equation}
and using (\ref{BayesRule})
\begin{eqnarray}\label{emmu}
\ \ \ \ \ \mu^{\prime}(x,y)&\!=
&\!\frac{\sum\limits_{x_1}\!P(Y_1,..,Y_N\Big|X_1=x_1, X_0=x,Y_0=y)P(X_1=x_1\Big| X_0=x,Y_0=y)\mu(x,y)}{P(Y_1,..,Y_N)} \!\!\!\!\! \!\\\nonumber
&\!=
&\!\frac{\sum\limits_{x_1}\beta_0(x_1,y)p_{x\rightarrow x_1}\mu(x,y)}{\sum\limits_{\xi}\alpha_N(\xi)}.
\end{eqnarray}
\begin{remark}
1) Different iterations of $p_{x\rightarrow {\xi}},\mu(x,y)$ will be used on the left 
and right hand sides of (\ref{empxx},\ref{emmu}).
The new estimates on the left are denoted $p^{\prime}_{x\rightarrow {\xi}},\mu^{\prime}(x,y)$.
Moreover, $\alpha_N$ also depends on (the earlier iteration of) $\mu$
so the equation is not linear.
It should be thought of as a Bayes' rule with the $\mu$ on the right
being a prior (to incorporating the observations with the
current set of parameters) and the one on the 
left being a posterior.\\
2) Setting a $p_{x\rightarrow {\xi}}=0$ or $\mu(x,y)=0$ will result in it 
staying zero for all updates.
This effectively removes this parameter from the EM optimization update
and should be avoided unless it is known that one of these should be $0$.\\
3) If there is no successive observations with $Y_n=y$ and $Y_{n+1}=y'$ in
the actual observation sequence, then all new estimates $q^{\prime}_{y\rightarrow y'}\left(x\right)$ will either be
set to $0$ or close to it.
They might not be exactly zero due to the first term in the numerator
of (\ref{emqyy}) where we could have an estimate of $Y_0=y$ and an observed
$Y_1=y'$.
\end{remark}

Naturally, our solution degenerates to the Baum{-}Welch algorithm in the HMM case.  However, the extra Markov component of MOM complicates this algorithm and 
its derivation.  
We start with $\alpha$, which is the most similar to HMM.
Here, we have by the joint Markov property and (\ref{alphadef}) that:
\begin{eqnarray}\displaystyle \label{alpharec}
&&\alpha _{n}\left(x\right)\\\nonumber
 & =  & P\left(Y_{1},...,Y_{n},X_{n}=x\right)  \\\nonumber
 & =  & \sum_{x_{n-1}}P\left(Y_{1},...,Y_{n},X_{n-1}=x_{n-1},X_{n}=x\right)  \\\nonumber
& =  & \sum_{x_{n-1}}P\left(Y_{1},...,Y_{n-1},X_{n-1}=x_{n-1}\right)P(X_{n}=x,Y_n\Big|Y_{1},...,Y_{n-1},X_{n-1}=x_{n-1})  \\\nonumber
 & =  & q_{Y_{n-1}\rightarrow Y_{n}}\left(x\right)\sum_{x_{n-1}}
\alpha_{n-1}(x_{n-1})p_{x_{n-1}\rightarrow x}\ , 
\end{eqnarray}
which can be solved forward for {$\displaystyle n=2,3,...,N-1,N $}, starting at 
\begin{center}
{\begin{tabular}{r@{ }c@{ }l}
{$\alpha _{1}\left(x_{1}\right) $} & {$= $} & {$\displaystyle \sum_{x_0}\! \sum_{y_0}\mu(x_0,y_0)\,p_{x_0\rightarrow x_1}\, q_{y_0\rightarrow Y_{1}}\left(x_1\right). $}
\end{tabular}}
\end{center}
Recall $\alpha_0=\mu$ is assigned differently.

Our iterative estimates for $p_{x\rightarrow x'}$, $q_{y\rightarrow y'}(x)$
and $\mu(x,y)$ also rely on the second (backward) recursion for $\beta_n$.
It also follows from the Markov property, our transition probabilities
and (\ref{Xdependsim}, \ref{Ydependsim}) that:
\begin{eqnarray}
\beta _{n}\left(x\right) \label{betarec} 
&\! =  &\! P\left(Y_{n+1},...,Y_{N}\Big|X_{n+1}=x,Y_{n}\right)  \\\nonumber
  &\! =  &\! \!P\!\left(Y_{n+2},...,Y_{N}\Big|X_{n+1}=x,Y_{n+1},Y_{n}\right)P\left(Y_{n+1}\Big|X_{n+1}=x,Y_{n}\right)  \\\nonumber
  &\! =  &\!P\!\left(\!Y_{n+2},...,Y_{N}\Big|X_{n+1}=x,Y_{n+1}\right)q_{Y_{n}\rightarrow Y_{n+1}}\left(x\right)  \\\nonumber
  &\! =  &\! \sum\limits_{x'\in E}\!P\!\left(\!Y_{n+2},...,Y_{N}\Big|X_{n+2}=x',X_{n+1}=x,Y_{n+1}\!\right)\\\nonumber
&&\ \ *P\!\left(\!X_{n+2}=x'\Big|X_{n+1}=x,Y_{n+1}\!\right)\!q_{Y_{n}\rightarrow Y_{n+1}}\left(x\right)  \\\nonumber
  &\! =  &\! \sum\limits_{x'} \beta_{n+1}(x')p_{x\rightarrow x'}q_{Y_{n}\rightarrow Y_{n+1}}\left(x\right), 
\end{eqnarray}
which can be solved backward for {$\displaystyle n=N-2,N-3,...,3,2,1,0 $}, starting from 
\[
\beta _{N-1}\left(x\right)=P(Y_N\Big| X_N=x,Y_{N-1}) = q_{Y_{N-1}\rightarrow Y_N}(x) .
\]
It is worth noting that when we use $\beta_n$ with $n=0$ we will have
$Y_0=y$ is some fixed value of interest not the missed observation
that we never see and we use the notation $\beta_0(x_0,y_0)$.
We only see $Y_1, Y_2,...,Y_N$.

We now have everything required for our algorithm, which is given in
Algorithm \ref{alg1} in Section \ref{EMSmall}.

To be able to show convergence in Section \ref{ConProb}, we need to track 
when parameters could become $0$.
The following lemma follows immediately from (\ref{alpharec}), 
(\ref{betarec}), induction and the fact that $\sum\limits_{x'}p_{x\rightarrow x'}=1$.
Any sensible initialization of our EM algorithm would ensure the condition 
$q_{Y_n\rightarrow Y_{n+1}}(x)>0$ holds.

\begin{lemma}
Suppose $q_{Y_n\rightarrow Y_{n+1}}(x)>0$ for all $x\in E$ and $n\in\{1,...,N-1\}$.
Then,
\begin{enumerate}
\item $\beta_m(x)>0$ for all $x\in E$ and $m\in\{1,...,N-1\}$.
\item $\beta_0(x,y)>0$ for any $x\in E, y\in O$ such that $q_{y\rightarrow Y_{1}}(x)>0$.
\item $\alpha_m(x)>0$ for all $x\in E$ and $m\in\{1,...,N\}$ if 
both\\ $\sum\limits_{x'}p_{x'\rightarrow x}>0$ and 
$\sum\limits_{y_0}\mu(x_0,y_0)q_{y_0\rightarrow Y_1}(x)>0$ for all $x,x_0\in E$.
\item $\alpha_0(x,y)>0$ if $\mu(x,y)>0$.
\end{enumerate}
\end{lemma}
Notice the condtion $\sum\limits_{x'}p_{x'\rightarrow x}>0$ for all $x$ says
that any hidden state can be reached from at least one other state
while $\sum\limits_{y_0}\mu(x_0,y_0)q_{y_0\rightarrow Y_1}(x)>0$ for all 
$x,x_0$ ensures that all the initial hidden states are meaningful.
The following result is the key to ensuring that our non-zero parameters stay
non-zero.
It follows from the prior lemma as well as (\ref{empxx},\ref{emqyy},\ref{emmu},\ref{betarec}).
\begin{lemma}\label{staypos}
Suppose $N\ge 2$, $q_{Y_n\rightarrow Y_{n+1}}(x)>0$ for all $x\in E$ and $n\in\{1,...,N-1\}$,
$\sum\limits_{x'}p_{x'\rightarrow x}>0$ for all $x\in E$ and 
$\sum\limits_{y_0}\mu(x_0,y_0)q_{y_0\rightarrow Y_1}(x)>0$ for all $x,x_0\in E$.
Then,
\begin{enumerate}
\item $p'_{x\rightarrow x'}>0$ if and only if $p_{x\rightarrow x'}>0$ for
any $x,x'$.
\item $q'_{y\rightarrow y'}(x)>0$ for all $x\in E$ if either
$\sum\limits_{n=1}^{N-1}1_{Y_n=y,Y_{n+1}=y'}>0$ or $\mu(\xi,y)1_{Y_1=y'}q_{y\rightarrow Y_1}(\xi)>0$ for all $\xi\in E$.
\item $\mu'(x,y)>0$ if $\mu(x,y)>0$ and $q_{y\rightarrow Y_1}(\xi)>0$ for all $\xi\in E$. $\mu'(x,y)=0$ if $q_{y\rightarrow Y_1}(\xi)=0$ for all $\xi\in E$.
\end{enumerate}
\end{lemma}
The algorithm; given explicitly in Section \ref{EMSmall}; starts with initial
estimates of all $p_{x\rightarrow x'}$, $q_{y\rightarrow y'}(x)$,
$\mu(x,y)$; say $p^1_{x\rightarrow x'}$, $q^1_{y\rightarrow y'}(x)$,
$\mu^1(x,y)$; and uses the formula for $p'_{x\rightarrow x'}$, $q'_{y\rightarrow y'}(x)$,
$\mu'(x,y)$ to refine these estimates successively to the
next estimates $p^2_{x\rightarrow x'},q^2_{y\rightarrow y'}(x),\mu^2(x,y)$; 
$p^3_{x\rightarrow x'},q^3_{y\rightarrow y'}(x),\mu^3(x,y)$; etc.
It is important to know that our estimates 
$\{p^k_{x\rightarrow x'},q^k_{y\rightarrow y'}(x),\mu^k(x,y)\}$
are getting better as $k\rightarrow\infty$.
Lemma \ref{staypos} will be used in some cases to ensure that an initially positive parameter stays
positive as $k$ increases, which important in our proofs to follow.

\section{EM Algorithm and Small Number Problem}\label{EMSmall}

The raw algorithm that we have considered hitherto computes
$\alpha_n$ and $\beta_n$ recursively.
By their definitions, 
\begin{eqnarray}
\alpha_n(x)&\!=&\!P(Y_1,...,Y_n, X_n=x)\\
\beta_n(x)&\!=&\!P(Y_{n+1},...,Y_N\Big| X_{n+1}=x,Y_n)
\end{eqnarray}
both can get extremely small when $N$ is large.
In this case, $\alpha_1(x)$ would be a reasonable number as it is just
a probability of the event $\{Y_1=Y_1,X_0=x\}$.
However, $\beta_1(x)$ would be a conditional probability of an exact
occurrence of $Y_2,...,Y_N$, which would usually be extraordinarily small.
Conversely, $\alpha_N(x)$ would usually be extraordinarily small and
$\beta_N(x)$ may be a reasonable number.
In between, the product $\alpha_n(x)\beta_n(x)$ would usually be 
extraordinarily small.
The unfortunate side-effect of this is that our $p_{x\rightarrow x'}$
(and $q$) calculations are basically going to result in zero over zero
most of the time when a computer is employed.
We need a fix.

This small number problem is resolved by using the filter instead of
$\alpha$.
Observe that the filter
\[
\pi_n(x)=P(X_n=x|Y_1,...,Y_n)=\frac{\alpha_n(x)}{\sum\limits_{\xi}\alpha_n(\xi)}
\]
is a (conditional) probability of a single event regardless
of $n$.
Hence, it does not necessarily get extraordinarily small.
However, scaling $\alpha_n$ in a manner depending upon $n$ means 
we will have to scale $\beta_n$ as well in a counteracting way.
The idea is to note that $\alpha_n(x)\beta_n(x)$ appear together
in computing the $p'_{x\rightarrow x'}$ and $q'_{y\rightarrow y'}(x)$
in such a way that we can divide every $\alpha_n(x)\beta_n(x)$ by
the same small number \emph{without changing} the values of the $p$'s and $q$'s.
Specifically, we replace 
\[
\alpha_n(x)\beta_n(x)\Rightarrow \frac{\alpha_n(x)\beta_n(x)}{a_1a_2\cdots
a_{N-1}}=\pi_n(x)\chi_n(x),\ \forall n\in\{1,...,N-1\},
\]
where $\pi_n(x)$ is the filter and $\chi_n(x)=\frac{\beta_n(x)}{a_{n+1}\cdots
a_{N-1}}$.
$a_1,...,a_N$ are normalizing constants and $\alpha_n(x,y)\beta_n(x,y)$
is scaled similarly.
Using (\ref{alpharec},\ref{betarec}), one finds the recursions for $\pi$ and $\chi$
are:
\begin{eqnarray}\displaystyle \label{pirec}
\rho _{n}\left(x\right)  & =  &  q_{Y_{n-1}\rightarrow Y_{n}}\left(x\right)\sum_{x_{n-1}}
\pi_{n-1}(x_{n-1})p_{x_{n-1}\rightarrow x}\ , \\\nonumber
\pi _{n}\left(x\right)  & =  & \frac{\rho _{n}\left(x\right)}{a_n},\ 
a_n=\sum\limits_{x_n}\rho_n(x_n),  
\end{eqnarray}
which can be solved forward for {$\displaystyle n=2,3,...,N-1,N $}, starting at 
\[\displaystyle 
\pi _{1}\left(x\right) = \frac{
\sum\limits_{x_0}\! \sum\limits_{y_0}\mu(x_0,y_0)\,p_{x_0\rightarrow x}\, q_{y_0\rightarrow Y_{1}}\left(x_1\right)}{a_1},\ a_1= 
\sum\limits_{x_1}\sum\limits_{x_0}\! \sum\limits_{y_0}\mu(x_0,y_0)\,p_{x_0\rightarrow x_1}\, q_{y_0\rightarrow Y_{1}}\left(x_1\right) . 
\]
Like $\beta$, $\chi$ is a backward recursion starting from
\[
\chi _{N-1}\left(x\right)=P(Y_N\Big| X_N=x,Y_{N-1}) = q_{Y_{N-1}\rightarrow Y_N}(x) 
\]
and then continuing as
\begin{eqnarray}
\chi _{n}\left(x\right) \label{chirec} 
&\! =   &\!\frac{q_{Y_{n}\rightarrow Y_{n+1}}\left(x\right) }{a_{n+1}}\sum\limits_{x'} {\chi_{n+1}(x')}p_{x\rightarrow x'}, 
\end{eqnarray}
which can be solved backward for {$\displaystyle n=N-2,N-3,...,3,2,1$}.
\begin{algorithm}[!ht]\label{alg1}
\DontPrintSemicolon
  \vspace*{0.2cm}
  \KwData{Observation sequence: $Y_1,...,Y_N$}
  \KwInput{Initial Estimates: $\{p_{x\rightarrow x'}\},\{q_{y\rightarrow y'}(x)\},\{\mu(x,y)\}$}
  \KwOutput{Final Estimates: $\{p_{x\rightarrow x'}\}$, $\{q_{y\rightarrow y'}(x)\}$, $\{\mu(x,y)\}$ \tcp*{Characterize MOM models}}
  \vspace*{0.2cm}
 \tcc{Initalization.}
  \While {$p$, $q$, and $\mu$ have not converged}{
  \tcc{Forward propagation.}
$\pi _{0}\left(x,y\right) = \mu(x,y)\ \forall x\in E,y\in O$;\\ 
$\rho _{1}\left(x\right) = \displaystyle \sum_{x_0\in E} \sum_{y_0\in O}\mu(x_0,y_0)\,p_{x_0\rightarrow x}\, q_{y_0\rightarrow Y_{1}}\left(x\right)\ \forall x\in E$;\\
$a_1 =\sum_{x}\rho_{1}\left(x\right)$\\
$\pi_{1}\left(x\right)=  \frac{\rho_{1}\left(x\right)}{a_1}$.\\
\For {$\displaystyle n=2,3,...,N $} {
$\displaystyle
\rho_{n}\left(x\right)  =  q_{Y_{n-1}\rightarrow Y_{n}}(x)\!\sum_{x_{n-1}\in E}\pi_{n-1}(x_{n-1})p_{x_{n-1}\rightarrow x} \ \forall x\in E.$\\
$a_n =\sum_{x}\rho_{n}\left(x\right)$.\\
$\pi_{n}\left(x\right)=  \frac{\rho_{n}\left(x\right)}{a_n}$.}
\tcc{Backward propagation.}
$\displaystyle\chi _{N-1}\left(x\right)= q_{Y_{N-1}\rightarrow Y_{N}}\left(x\right) \ \forall x\in E$.\\
\For {$\displaystyle n=N-2,N-3,...,1 $} {
$\chi _{n}\left(x\right)=\frac{q_{Y_{n}\rightarrow Y_{n+1}}\!\left(x\right)}{a_{n+1}} \sum\limits_{x'\in E}\chi_{n+1}(x')p_{x\rightarrow x'}\ \forall x\in E$.
}
$\chi _{0}\left(x,y\right)=\frac{q_{y\rightarrow Y_{1}}\!\left(x\right)}{a_1} \sum\limits_{x'\in E}\chi_{1}(x')p_{x\rightarrow x'}\ \forall x\in E, y\in O$.\\
\tcc{Probability Update.}
$q_{y\rightarrow y'}\!\left(x\right) =  \displaystyle \frac{\sum\limits_{\xi}p_{\xi\rightarrow x}\left[1_{Y_{1}=y'}\chi_{0}(x,y)\pi_{0}\left(\xi,y\right)+\sum\limits _{n=1}^{N-1}\!1_{Y_{n}=y,Y_{n+1}=y'}\chi _{n}\!\left(x\right)\pi_{n}(\xi)\right]}{\sum\limits_{\xi}p_{\xi\rightarrow x}\left[\chi_{0}(x,y)\pi_{0}\left(\xi,y\right)+\sum\limits _{n=1}^{N-1}\!1_{Y_{n}=y}\chi _{n}\!\left(x\right)\pi_{n}(\xi)\right]}$ \\ $\forall x\in E; y,y'\in O$.\\
$\displaystyle \mu \left(x,y\right)=\frac{\mu\left(x,y\right)\sum_{x_1} \chi_0(x_1,y)p_{x\rightarrow x_1}}{\sum_\xi\sum_\theta \mu\left(\xi,\theta\right)\sum_{x_1} \chi_0(x_1,\theta)p_{\xi\rightarrow x_1}} \ \forall x\in E; y\in O.$\\
$\displaystyle
p_{x\rightarrow {x^{\prime}}} =\frac{p_{x\rightarrow {x^{\prime}}}\left[\sum\limits_y\pi_0(x,y)\chi_0(x^{\prime},y)+\sum\limits _{n=1}^{N-1}\pi_n(x)\chi_n(x^{\prime})\right]}{\sum\limits_{x_1}p_{x\rightarrow {x_1}}\left[\sum\limits_y\pi_0(x,y)\chi_0(x_1,y)+\sum\limits _{n=1}^{N-1}\chi _{n}\left(x_1\right)\pi_n(x)\right]}  \ \forall x,x'\in E$. 
}
\caption{EM algorithm for MOM}
\end{algorithm}

Finally, the $n=0$ value for $\pi$ and $\chi$ become
\begin{eqnarray}
\chi _{0}\left(x,y\right) \label{chizero} 
&\! =   &\! \sum\limits_{x'} \frac{\chi_{1}(x')}{a_{1}}p_{x\rightarrow x'}q_{y\rightarrow Y_{1}}\left(x\right), \\
\pi _{0}\left(x,y\right) \label{pizero} 
&\! =   &\! \alpha _{0}\left(x,y\right)=\mu\left(x,y\right).
\end{eqnarray}
The adjusted, non-raw algorithm is given in Algorithm \ref{alg1}

{\bf Note:} In the three probability ($p,q,\mu$) update steps of Algorithm \ref{alg1}, 
it usually better from numeric
and performance perspectives to compute the numerators and then use
the facts that they must be probability mass functions to properly
normalize rather than use the full equation as given.

\section{Convergence of Probabilities}\label{ConProb}

In this section, we establish the convergence properties of the transition
probabilities and initial distribution
$\{p^k_{x\rightarrow x'},q^k_{y\rightarrow y'}(x),\mu^k(x,y)\}$ that we derived in Section \ref{ProbEM}.
Our method adapts the ideas of Baum et. al. \cite{Baum3}, Liporace 
\cite{Liporace} and Wu \cite{Wu} to our setting.

We think of the transition probabilities and initial distribution as parameters, 
and let $\Theta$ denote all of the \emph{non-zero} transition and 
initial distribution probabilities in $p, q, \mu$.
Let $e=|E|$ and $o=|O|$ be the cardinalities of the hidden and
observation spaces.
Then, the whole parameter space has cardinality $d'=e^2+e*o^2+e*o$ for the
$p_{x\rightarrow x'}$ plus $q_{y\rightarrow y'}(x)$ plus $\mu(x,y)$ and
lives on $[0,1]^{d'}$.
However, we are removing the values that will be set to zero and adding
\emph{sum to one} constraints to consider a constrained optimization problem
on $(0,\infty)^d$ for some $d\le d'$.
Removing these zero possibilities gives us necessary regularity
for our re-estimation procedure.
However, it was not enough to just remove them at the beginning.
We had to ensure that zero parameters did not creep in during our
interations or else we will be doing such things as taking logarithms
of $0$.
Lemma \ref{staypos} suggests a strategy for initially assigning estimates
so zeros will not occur in later estimates in the case
that the value of $Y_1$ also appears later in the observation sequence.
\begin{enumerate}
\item Pick initial estimate $\{p^1_{x\rightarrow x'}\}$ such that $\sum\limits_{x}p^1_{x\rightarrow x'}>0$ for all $x'$.  
This says that any hidden state can be reached from somewhere.
From above we know $p^1_{x\rightarrow x'}\rightarrow p^k_{x\rightarrow x'}$
for all $k$ so $\sum\limits_{x}p^k_{x\rightarrow x'}>0$ for all $x'$.  
\item Pick $q^1_{y\rightarrow y'}(x)>0$ for all $x\in E$ if and only if $\sum\limits_{n=1}^{N-1}1_{Y_n=y,Y_{n+1}=y'}>0$.
\item Pick $\mu^1(x,y)>0$ if and only if $q^1_{y\rightarrow Y_1}(x)>0$.
Here you are using the values you just picked in the previous step
to make this decision.
\end{enumerate}
This will produce an example of a \emph{zero separating} sequence
in the case the value of $Y_1$ is repeated as at least one $Y_n$ with $n>1$.

\begin{definition}
A sequence of estimates $\{p^k,q^k,\mu^k\}$ is zero separating if:
\begin{enumerate}
\item $p^1_{x\rightarrow x'}>0$ iff $p^{k}_{x\rightarrow x'}>0$ for all $k=1,2,3,...$,
\item $q^{1}_{y\rightarrow y'}(x)>0$ iff
$q^{k}_{y\rightarrow y'}(x)>0$ for all $k=1,2,3,...$, and
\item $\mu^1(x,y)>0$ iff $\mu^{k}(x,y)>0$ for all $k=1,2,3,...$.
\end{enumerate}
Here, iff stands for if and only if.
\end{definition}
This means that we can potentially optimize over $p,q,\mu$ that we initially
do not set to zero.
Henceforth, we factor the zero $p,\mu, q$ out of $\Theta$, consider
$\Theta\subset (0,\infty)^d$ with $d\le d'$ and 
define the parameterized mass functions
\begin{eqnarray}
\!&&p_{y_0,y_1,...,y_N}(x;\!\Theta\!)\\\nonumber
&\!=& 
p_{x_0\rightarrow x_1}q_{y_0\rightarrow y_1}(x_1)
p_{x_1\rightarrow x_2}q_{y_1\rightarrow y_2}(x_2)\cdots
p_{x_{N-1}\rightarrow x_N}q_{y_{N-1}\rightarrow y_N}(x_N\!)\mu(x_0,y_0)
\end{eqnarray}	
in terms of the \emph{non-zero} values only.
The observable likelihood
\begin{eqnarray}
P_{Y_1,...,Y_N}(\Theta)
&\!=&\!\! \sum_{x_0,x_1,...,x_N}\sum_{y_0}
p_{y_0,Y_1,...,Y_N}(x_0,x_1,...,x_N;\Theta)
\end{eqnarray}	
is not changed by removing the zero values of $p,\mu,q$
and this removal allows us to define the re-estimation function
\begin{eqnarray}
\ \ \ \ Q_{Y_1,...,Y_N}(\Theta,\Theta')
&=&\! \sum_{x_0,...,x_N}\sum_{y_0}
p_{y_0,Y_1,...,Y_N}(x_0,...,x_N;\Theta)\ln p_{y_0,Y_1,...,Y_N}(x_0,...,x_N;\Theta').\!\!\!
\end{eqnarray}	
{\bf Note:} Here and in the sequel, 
the summation in $P,Q$ above are only over the non-zero
combinations.
We would not include an $x_i, x_{i+1}$ pair where $p_{x_i\rightarrow x_{i+1}}=0$
nor an $x_0, y_0$ pair where $\mu(x_0,y_0)=0$.
Hence, our parameter space is
\[
\Gamma=\{\Theta\in (0,\infty)^d: \sum\limits_{x'}p_{x\rightarrow x'}=1,
 \sum\limits_{y'}q_{y\rightarrow y'}(x)=1\ \forall x, 
\sum\limits_{x,y}\mu(x,y)=1
\}.
\]
Later, we will consider the extended parameter space 
\[
K=\{\Theta\in [0,1]^d: \sum\limits_{x'}p_{x\rightarrow x'}=1,
 \sum\limits_{y'}q_{y\rightarrow y'}(x)=1\ \forall x, 
\sum\limits_{x,y}\mu(x,y)=1
\}
\]
as limit points.
Note: In both $\Gamma$ and $K$, $\Theta$ is only over the 
$p_{x\rightarrow x'}$, $q_{y\rightarrow y'}(x)$ and
$\mu(x,y)$ that are not just set to $0$ (before limits).

Then, equating $Y_0$ with $y_0$ to ease notation, one has that
\begin{eqnarray}\label{Qformula}
Q(\Theta,\Theta')
&\!=&\!\! \sum_{x_0,...,x_N}\!\sum_{y_0}
\left[\prod_{n=1}^Np_{x_{n-1}\rightarrow x_n}q_{Y_{n-1}\rightarrow Y_n}(x_n)\right]\mu(x_0,y_0)\\\nonumber
&\!\!\!\!&\!\!\!\left[\sum_{m=1}^N\!\left\{\ln p'_{x_{m-1}\rightarrow x_m}\!+\ln q'_{Y_{m-1}\rightarrow Y_m}(x_m)\right\}+\ln\mu'(x_0,y_0)\right].\ \ \
\end{eqnarray}
The re-estimation function will be used to interpret
the EM algorithm we derived earlier.
We impose the following condition to ensure everything is well defined.
\begin{description}
\item[(Zero)] The EM estimates are zero separating.
\end{description}
The following result that is motivated by Theorem
3 of Liporace \cite{Liporace}.
\begin{theorem}
Suppose (Zero) holds.  The expectation-maximization solutions (\ref{empxx}, \ref{emqyy}, \ref{emmu}) 
derived in Section \ref{ProbEM} are the \emph{unique} critical
point of the re-estimation function $\Theta'\rightarrow Q(\Theta,\Theta')$,
subject to $\Theta'$ forming probability mass functions.
This critical point is a maximum taking value in $(0,1]^d$ for $d$
explained above.
\end{theorem}
We consider it as an optimization problem over the open set $(0,\infty)^d$ but 
with the constraint that we have mass functions so the values have to be
in the set $(0,1]^d$.\\
\proof
One has by (\ref{Qformula}) as well as the constraint 
$\sum_{x'}p'_{x\rightarrow x'}=1$ that the maximum must satisfy
\begin{eqnarray}\label{derivx}
0&\!= &\!\frac{\partial}{\partial
p'_{x\rightarrow x'}}\left\{ Q(\Theta,\Theta')-\lambda\left(\sum_{\xi}
p'_{x\rightarrow \xi}-1\right)\right\}\\\nonumber
&\!= &\!\sum_{x_0,...,x_N}\!\sum_{y_0}
\left[\prod_{n=1}^Np_{x_{n-1}\rightarrow x_n}q_{Y_{n-1}\rightarrow Y_n}(x_n)\right]
\!\sum_{m=1}^N\frac{1_{x_{m-1}=x}1_{x_{m}=x'}}{p'_{x\rightarrow x'}}\mu(x_0,y_0)-\lambda
\end{eqnarray}
where 
$\lambda$ is a Lagrange multiplier.
Multiplying by $p'_{x\rightarrow x'}$, summing over $x'$ and then
using the Markov property as well as the argument in (\ref{XnXnm1},\ref{Xn}), 
one has that
\begin{eqnarray}\label{lambdamx}
\lambda&\!= &\!\sum_{m=1}^N\sum_{x_0,...,x_N}\!\sum_{y_0}
\left[\prod_{n=1}^Np_{x_{n-1}\rightarrow x_n}q_{Y_{n-1}\rightarrow Y_n}(x_n)\right]
\!1_{x_{m-1}=x}\,\mu(x_0,y_0)\\\nonumber
&\!= &\!\sum_{m=1}^N P(X_{m-1}=x,Y_1,...,Y_N)\\\nonumber
&\!= &\!\sum_{y}\sum_{x_1}\beta_{0}(x_1,y)p_{x\rightarrow x_1}\alpha_{0}(x,y)+\sum_{m=2}^N\sum_{x_m}\beta_{m-1}(x_m)p_{x\rightarrow x_m}\alpha_{m-1}(x).
\end{eqnarray}
Substituting (\ref{lambdamx}) into (\ref{derivx}), one has by the
Markov property that
\begin{eqnarray}\label{finix}\displaystyle
p'_{x\rightarrow x'}&\!= &\!\sum\limits_{x_0,...,x_N}\!\sum\limits_{y_0}
\left[\prod\limits_{n=1}^Np_{x_{n-1}\rightarrow x_n}q_{Y_{n-1}\rightarrow Y_n}(x_n)\right]
\!\sum\limits_{m=1}^N\frac{1_{x_{m-1}=x}1_{x_{m}=x'}}{\lambda}\mu(x_0,y_0)\\\nonumber
&\!= &\!\frac{\sum\limits_{m=1}^NP(X_{m-1}=x,X_m=x',Y_1,...,Y_N)}{\sum\limits_{y}\sum\limits_{x_1}\beta_{0}(x_1,y)p_{x\rightarrow x_1}\alpha_{0}(x,y)+\sum\limits_{m=2}^N\sum\limits_{x_m}\beta_{m-1}(x_m)p_{x\rightarrow x_m}\alpha_{m-1}(x)}\\\nonumber
&\!= &\!\frac{\sum\limits_{y}\beta_{0}(x',y)p_{x\rightarrow x'}\alpha_{0}(x,y)+\sum\limits_{m=2}^N\beta_{m-1}(x')p_{x\rightarrow x'}\alpha_{m-1}(x)}{\sum\limits_{y}\sum\limits_{x_1}\beta_{0}(x_1,y)p_{x\rightarrow x_1}\alpha_{0}(x,y)+\sum\limits_{m=2}^N\sum\limits_{x_m}\beta_{m-1}(x_m)p_{x\rightarrow x_m}\alpha_{m-1}(x)}.
\end{eqnarray}
Clearly, the value on the far right of (\ref{finix}) is in $(0,1]$ (since
we assumed $p_{x\rightarrow x'}>0$).
Similarly,
\begin{eqnarray}\label{derivy}
\ \ \ \ 0&\!= &\!\frac{\partial}{\partial
q'_{y\rightarrow y'}(x)}\left\{ Q(\Theta,\Theta')-\lambda\left(\sum_{\theta\in O}
q'_{y\rightarrow \theta}(x)-1\right)\right\}\\\nonumber
&\!= &\!\!\sum_{x_0,...,x_N}\!\sum_{y_0}
\left[\prod_{n=1}^Np_{x_{n-1}\rightarrow x_n}q_{Y_{n-1}\rightarrow Y_n}(x_n)\right]
\!\sum_{m=1}^N\frac{1_{Y_{m-1}=y}1_{Y_{m}=y'}1_{X_{m}=x}}{q'_{y\rightarrow y'}(x)}\mu(x_0,y_0)-\lambda,\!\!\!
\end{eqnarray}
where 
$\lambda$ is a Lagrange multiplier.
Multiplying by $q'_{y\rightarrow y'}(x)$, summing over $y'$ and then
using the Markov property as well as the argument in (\ref{XnXnm1},\ref{Xn2}), one has that
\begin{eqnarray}\label{lambdamy}
\lambda&\!= &\!\sum_{m=1}^N\sum_{x_0,...,x_N}\!\sum_{y_0}
\left[\prod_{n=1}^Np_{x_{n-1}\rightarrow x_n}q_{Y_{n-1}\rightarrow Y_n}(x_n)\right]
\!1_{Y_{m-1}=y}1_{x_{m}=x}\mu(x_0,y_0)\\\nonumber
&\!= &\!P(Y_0=y,X_{1}=x,Y_1,...,Y_N)+\sum_{m=2}^N 1_{Y_{m-1}=y}P(X_{m}=x,Y_1,...,Y_N)\\\nonumber
&\!= &\!\beta_0(x,y)\sum_{x_0}p_{x_0\rightarrow x}\mu(x_0,y)+\sum_{n=1}^{N-1}1_{Y_{n}=y}\beta_{n}(x)\sum_{x_{n}}p_{x_{n}\rightarrow x}\alpha_{n}(x_n).
\end{eqnarray}
Substituting (\ref{lambdamy}) into (\ref{derivy}), one has that
\begin{eqnarray}\label{finiy}\displaystyle
&\! &\!q'_{y\rightarrow y'}(x)\\\nonumber
&\!= &\!
\frac{P(Y_0=y,X_{1}=x,Y_1=y',Y_2,...,Y_N)+\!\sum\limits_{m=2}^N\!1_{Y_{m-1}=y,Y_{m}=y'}P(X_m\!=x,Y_1,...,Y_N)}{\beta_0(x,y)\sum\limits_{x_0}p_{x_0\rightarrow x}\mu(x_0,y)+\sum\limits_{n=1}^{N-1}1_{Y_{n}=y}\beta_{n}(x)\sum\limits_{x_{n}}p_{x_{n}\rightarrow x}\alpha_{n}(x_n)}\\\nonumber
&\!= &\!\frac{1_{Y_1=y'}\beta_0(x,y)\sum\limits_{x_0}p_{x_0\rightarrow x}\mu(x_0,y)+\sum\limits_{n=1}^{N-1}1_{Y_{n}=y}1_{Y_{n+1}=y'}\beta_{n}(x)\sum\limits_{x_{n}}p_{x_{n}\rightarrow x}\alpha_{n}(x_n)}{\beta_0(x,y)\sum\limits_{x_0}p_{x_0\rightarrow x}\mu(x_0,y)+\sum\limits_{n=1}^{N-1}1_{Y_{n}=y}\beta_{n}(x)\sum\limits_{x_{n}}p_{x_{n}\rightarrow x}\alpha_{n}(x_n)}.
\end{eqnarray}
Finally, for a maximum one also requires
\begin{eqnarray}\label{derivmu}
0&\!= &\!\frac{\partial}{\partial
\mu'(x,y)}\left\{ Q(\Theta,\Theta')-\lambda\left(\sum_{\xi\in E,\theta\in O}
\mu'(\xi,\theta)-1\right)\right\}\\\nonumber
&\!= &\!\sum_{x_0,...,x_N}\!\sum_{y_0}
\left[\prod_{n=1}^Np_{x_{n-1}\rightarrow x_n}q_{Y_{n-1}\rightarrow Y_n}(x_n)\right]
\!\frac{1_{x_0=x}1_{y_0=y}}{\mu'(x,y)}\mu(x_0,y_0)-\lambda,
\end{eqnarray}
where 
$\lambda$ is a Lagrange multiplier.
Multiplying by $\mu'(x,y)$ and summing over $x,y$, one has that
\begin{eqnarray}\label{lambdammu}
\lambda&= &\!\sum\limits_{x_0,...,x_N}\!\sum\limits_{y_0}
\left[\prod\limits_{n=1}^Np_{x_{n-1}\rightarrow x_n}q_{Y_{n-1}\rightarrow Y_n}(x_n)\right]
\!\mu(x_0,y_0)\\\nonumber
&= &\!P(Y_1,...,Y_N)\\\nonumber
&= &\!\sum_{\xi}\alpha_{N}(\xi).
\end{eqnarray}
Substituting (\ref{lambdammu}) into (\ref{derivmu}), one has by 
(\ref{Xdependsim},\ref{Ydependsim}) that
\begin{eqnarray}\label{finimu}\displaystyle
\ \ \ \ \ \ \mu'(x,y)&\!= &\!
\frac{\sum\limits_{x_0,...,x_N}\!\sum\limits_{y_0}
\left[\prod\limits_{n=1}^Np_{x_{n-1}\rightarrow x_n}q_{Y_{n-1}\rightarrow Y_n}(x_n)\right]
\!1_{x_0=x}1_{y_0=y}\mu(x_0,y_0)}{\sum\limits_{\xi}\alpha_{N}(\xi)}
\\\nonumber
&\!= &\!\frac{P(X_0=x,Y_0=y,Y_1,...,Y_N)}{\sum\limits_{\xi}\alpha_{N}(\xi)}
\\\nonumber
&\!= &\!\frac{\sum\limits_{x_1}P(Y_1,...,Y_N|X_1=x_1,X_0=x,Y_0=y)P(X_1=x_1|X_0=x,Y_0=y)\mu(x,y)}{\sum\limits_{\xi}\alpha_{N}(\xi)}\!\!\!\!\!
\\\nonumber
&\!= &\!\frac{\sum\limits_{x_1}\beta_{0}(x_1,y)p_{x\rightarrow x_1}\mu(x,y)}{\sum\limits_{\xi}\alpha_{N}(\xi)}.
\end{eqnarray}
If we were to sum the numerator on the far right of (\ref{finimu}), then
upon substitution of $\beta$ we would get $P(Y_1,...,Y_N)$, which matches
the denominator.
Hence, $\mu'(x,y)\in [0,1]$ like the other new estimates.
Now, we have established that the EM algorithm of Section \ref{ProbEM}
corresponds to the unique critical point of $\Theta'\rightarrow Q(\Theta,\Theta')$.
Moreover, all mixed partial derivative of $Q$ in the components of $\Theta'$
are $0$, while
\begin{eqnarray}
&\!\!\!&\!\frac{\partial^2 Q_{Y_1,Y_2,...,Y_N}(\Theta,\Theta')}{\partial
{p^\prime_{x\rightarrow x'}}^2}\\\nonumber
&\!\!\!=&\!-\!\!\sum_{y_0;x_0,...,x_N}
\!\left[\prod_{n=1}^Np_{x_{n-1}\rightarrow x_n}q_{Y_{n-1}\rightarrow Y_n}(x_n)\right]\sum\limits_{m=1}^N \frac{1_{x_{m-1}=x,x_{m}=x'}}{{p^\prime_{x\rightarrow x'}}^2}
\mu(x_0,y_0)\ \ \ \ 
\end{eqnarray}
\begin{eqnarray}
&\!\!\!&\!\frac{\partial^2 Q_{Y_1,Y_2,...,Y_N}(\Theta,\Theta')}{
{\partial q^\prime_{y\rightarrow y'}(x)}^2}\\\nonumber
&\!\!\!=&\!-\!\sum_{y_0;x_0,...,x_N}
\!\left[\prod_{n=1}^N\!p_{x_{n-1}\rightarrow x_n}q_{Y_{n-1}\rightarrow Y_n}(x_n)\right]\sum\limits_{m=1}^N\! \frac{1_{Y_{m-1}=y,Y_m=y',x_m=x}}{{ q^\prime_{y\rightarrow y'}(x)}^2}
\mu(x_0,y_0)\ \ \ \ \ \
\end{eqnarray}
and
\begin{eqnarray}
&\!\!\!&\!\frac{\partial^2 Q_{Y_1,Y_2,...,Y_N}(\Theta,\Theta')}{
{\partial \mu^\prime(x,y)}^2}\\\nonumber
&\!\!\!=&\!-\!\sum_{y_0;x_0,...,x_N}
\!\left[\prod_{n=1}^Np_{x_{n-1}\rightarrow x_n}q_{Y_{n-1}\rightarrow Y_n}(x_n)\!\right]\!\sum\limits_{m=1}^N\! \frac{1_{y_{0}=y,x_0=x}}{{\mu^\prime(x,y)}^2}
\mu(x_0,y_0).\ \ \ \ \
\end{eqnarray}
Hence, the Hessian matrix is diagonal with negative values along its axis
and the critical point is a maximum.
\endproof
The upshot of this result is that, if the EM algorithm produces parameters
$\{\Theta^k\}\subset \Gamma$, then $Q(\Theta^k,\Theta^{k+1})\ge Q(\Theta^k,\Theta^k)$.
Now, we have the following result, based upon 
Theorem 2.1 of Baum et.\ al.\ \cite{Baum3}, that establishes the observable
likelihood is also increasing i.e.\ $P(\Theta^{k+1})\ge P(\Theta^k)$.

\begin{lemma}\label{lemma1}
Suppose (Zero) holds.  
$Q(\Theta,\Theta')\ge Q(\Theta,\Theta)$ implies $P(\Theta')\ge P(\Theta)$.
Moreover, $Q(\Theta,\Theta')> Q(\Theta,\Theta)$ implies 
$P(\Theta')> P(\Theta)$.
\end{lemma}
\proof
$\ln (t)$ for $t>0$ has convex inverse $\exp (t)$.
Hence, by Jensen's inequality
\begin{eqnarray}\displaystyle
&\!&\ \ \ \ \ \ \frac{
Q(\Theta,\Theta')-Q(\Theta,\Theta)}{P(\Theta)}\\\nonumber
&\!\!=&\ln\exp\!\left[\sum_{x_0,x_1,...,x_N}\sum_{y_0}\ln\left(\frac{p_{y_0,Y_1,...,Y_N}(x_0,x_1,...,x_N;\Theta')}{p_{y_0,Y_1,...,Y_N}(x_0,x_1,...,x_N;\Theta)}\right)\frac{
p_{y_0,Y_1,...,Y_N}(x_0,x_1,...,x_N;\Theta)}{P(\Theta)}\right]\\\nonumber
&\!\le &
\ln\left(\frac{\sum\limits_{x_0,x_1,...,x_N}\sum\limits_{y_0}
p_{y_0,Y_1,...,Y_N}(x_0,x_1,...,x_N;\Theta)\frac{p_{y_0,Y_1,...,Y_N}(x_0,x_1,...,x_N;\Theta')}{p_{y_0,Y_1,...,Y_N}(x_0,x_1,...,x_N;\Theta)}}{P(\Theta)}\right)\\\nonumber
&\!= &\ln\left(\frac{P(\Theta')}{P(\Theta)}\right)
\end{eqnarray}
and the result follows.
\endproof

The stationary points of $P$ and $Q$ are also related.
\begin{lemma}
Suppose (Zero) holds.  
A point $\Theta\in \Gamma$ 
is a critical point of $P(\Theta)$ if and only if it is
a fixed point of the re-estimation function, i.e. 
$Q(\Theta; \Theta) = \max_{\Theta'} Q(\Theta;\Theta')$ since $Q$ is
differentiable on $(0,\infty)^d$ in $\Theta'$.
\end{lemma}
\proof
The following derivatives are equal:
\begin{eqnarray}
\ \ \ \ \ \frac{\partial P_{Y_1,...,Y_N}(\Theta)}{\partial p_{x\rightarrow x'}}
&\!=&\!\!\sum_{x_0,...,x_N}\!\sum_{y_0}\!
\left[\prod_{n=1}^Np_{x_{n-1}\rightarrow x_n}q_{Y_{n-1}\rightarrow Y_n}(x_n)\right]\!\sum\limits_{m=1}^N\! \frac{1_{x_{m-1}=x,x_{m}=x'}}{p_{x_{m-1}\rightarrow x_m}}
\mu(x_0,y_0)\!\!\\\nonumber
&\!=&\frac{\partial Q_{Y_1,Y_2,...,Y_N}(\Theta,\Theta')}{\partial
p'_{x\rightarrow x'}}\Big|_{\Theta'=\Theta},
\end{eqnarray}
which are defined since $p_{x\rightarrow x'}\ne 0$. Similarly,
\begin{eqnarray}\displaystyle
\ \ \ \ \ \ \frac{\partial P_{Y_1,...,Y_N}(\Theta)}{\partial q_{y\rightarrow y'}(x)}&\!=&\!\!\!\sum_{x_0,...,x_N}\!\!\sum_{y_0}
\!\left[\prod_{n=1}^N\!p_{x_{n-1}\rightarrow x_n}q_{Y_{n-1}\rightarrow Y_n}(x_n)\!\right]\!\sum\limits_{m=1}^N \!\frac{1_{Y_{m-1}=y,Y_m=y',x_m=x}}{q_{Y_{m-1}\rightarrow Y_m}(x)}
\mu(x_0,y_0)\!\!\!\!\!\!\!\!\!\!\\\nonumber
&\!\!\!=&\!\frac{\partial Q_{Y_1,Y_2,...,Y_N}(\Theta,\Theta')}{\partial
q'_{y\rightarrow y'}(x)}\Big|_{\Theta'=\Theta}
\end{eqnarray}
and
\begin{eqnarray}\displaystyle
\frac{\partial P_{Y_1,...,Y_N}(\Theta)}{\partial \mu(x,y)}&\!=&\!\sum_{x_0,...,x_N}\!\sum_{y_0}
\left[\prod_{n=1}^Np_{x_{n-1}\rightarrow x_n}q_{Y_{n-1}\rightarrow Y_n}(x_n)\right]
1_{(x_0,y_0)=(x,y)}\\\nonumber
&\!=&\!\frac{\partial Q_{Y_1,Y_2,...,Y_N}(\Theta,\Theta')}{\partial
\mu'(x,y)}\Big|_{\Theta'=\Theta}.
\end{eqnarray}
\endproof
We can rewrite (\ref{finix},\ref{finiy},\ref{finimu}) in recursive form with
the values of $\alpha$ and $\beta$ substituted in to find
that
\[
\Theta^{k+1}=M(\Theta^k),
\] 
where $M$ is a continuous function.
Moreover, $P:K\rightarrow [0,1]$ is continuous and satisfies
$P(\Theta^k)\le P(M(\Theta^k))$ from above.
Now, we have established everything we need for the following result,
which follows from the proof of Theorem 1 of \cite{Wu}.

\begin{theorem}
Suppose (Zero) holds.  
Then, $\{\Theta^k\}_{k=1}^\infty$ is relatively compact, all its limit points 
(in $K$) are stationary points of $P$, producing the same likelihood $P(\Theta^*)$ say, 
and $P(\Theta^k)$ converges monotonically to $P(\Theta^*)$. 
\end{theorem}

\cite{Wu} has several interesting results in the context of general
EM algorithms to guarantee convergence to local or global maxima
under certain conditions.
However, the point of this note is to introduce a new model and
algorithms with just enough theory to justify the algorithms.
Hence, we do not consider theory under any special cases here
but rather refer the reader to Wu \cite{Wu}.

\section{Viterbi algorithm}\label{Viterbi}

Like for HMM, the filter for MOM can be computed in real time (once the parameters are known)
\begin{center}
{\begin{tabular}{r@{ }c@{ }l}
{$\pi _{n}\left(x\right) $} & {$= $} & {$P\left(X_{n}=x\Big|Y_{1},...,Y_{n}\right) $}, {$\ \forall x\in E. $}
\end{tabular}}
\end{center}So, we can just compute {$\displaystyle \alpha _{n}\left(x\right) $} and then just normalize i.e. 
\begin{equation}\label{filter}
\displaystyle \pi _{n}\left(x\right)=\frac{\alpha _{n}\left(x\right)}{\sum\limits _{\xi }\alpha _{n}\left(\xi \right)}\ \mbox{ and }\pi(A)=\sum_{x\in A}\pi(x). 
\end{equation}
This provides our tracking estimate of the hidden state given the observations.  Prediction can then be done by running the Kolmogorov forward equation starting from this estimate.

We can compute the most likely single hidden state values $x^+_n$ 
of the hidden state $X_n$, given the back observations $Y_1,...,Y_n$ 
by finding the values that maximize 
$x\rightarrow P(X_n=x\Big|Y_{1},...,Y_{n})$.
However, the Viterbi algorithm is used in HMM to find the most likely
whole sequence of hidden state given the complete sequence of observations.
This is particularly important in problems like decoding or recognition
but is still useful in a widerange of application.
It is a dynamic programming type algorithm.

As there is a EM analog to the Baum-Welch algorithm for our MOM models,
it is natural to wonder if there is a dynamic programming analog to
the Viterbi algorithm for our MOM models.
The answer is yes and it is more similar to the Viterbi algorithm
than our MOM EM algorithm is to the Baum-Welch algorithm.
There are three small variants that one can consider: finding
the most likely sequence including both the initial hidden state $X_0$ and
the unseen observation $Y_0$, including just the hidden state $X_0$ or 
neither.  We consider doing both $X_0$ and $Y_0$ here.
(The others are basically the same, starting with a marginal of 
our initial distribution given in our algorithm here.)

We define a sequence of functions $\delta_{0,1}(y,x_0,x_1),\{\delta_n(x)\}_{n=2}^N$, the maximum functions, and a sequence of
estimates $\{y^*_0,x^*_0,x^*_1,...,x^*_N\}$, the most likely sequence, within our Viterbi algorithm, Algorithm
\ref{alg2} below. 
Then, we show the algorithm works by noting
\begin{equation}\label{deltaver}
\delta_n(x)=\max_{y_0;x_0,x_1,...,x_{n-1}}\!\!P(Y_0=y_0;X_0=x_0,...,X_{n-1}=x_{n-1};X_n=x;Y_1,...,Y_n),
\end{equation}
for all $n,x$ and establishing that $\{y^*_0,x^*_0,...,x^*_N\}$ satisfies $x^*_N=\mathop{\arg\max}\limits_x\delta_N(x)$ and 
$$
\delta_N(x^*_N)=P(Y_0=y^*_0;X_0=x^*_0,X_1=x^*_1,...,X_{N-1}=x^*_{N-1};X_N=x^*_N;Y_1,...,Y_N).
$$

\begin{algorithm}[!ht]\label{alg2}
\DontPrintSemicolon
  \vspace*{0.2cm}
  \KwInput{Observation sequence: $Y_1,...,Y_N$}
  \KwOutput{Most likely Hidden state sequence: $P^*$; $y^*_0;x^*_0,x^*_1,...,x^*_N$}
  \KwData{Probabilities $\{p_{x\rightarrow x'}\}$, $\{q_{y\rightarrow y'}(x)\}$, $\{\mu(x,y)\}$ \tcp*{Distinguish MOM models}}
  \vspace*{0.2cm}
  $\delta_{0,1}(y_0,x_0,x_1)=\mu(x_0,y_0)p_{x_0\rightarrow x_1}q_{y_0\rightarrow Y_1}(x_1),\ \forall y_0,x_0,x_1$ \tcp*{Initialize joint distribution.}
  \vspace*{0.2cm}
  \tcc{Substitute Marginal $\delta_{1}(x_1)=\sum\limits_{y_0,x_0}\delta_{0,1}(y_0,x_0,x_1)$ if want $x^*_1,...,x^*_N$.}
$\delta_2(x_2)=\max\limits_{y_0\in O;x_0,x_1\in E}\left[\delta_{0,1}(y_0,x_0,x_1)p_{x_{1}\rightarrow x_2}\right]q_{Y_{1}\rightarrow Y_2}(x_2)$\\
  \tcc{Replace $\delta_n$ with normalized $\gamma_n$ given below to
avoid small number problem.}
 $\psi_2(x_2)=\mathop{\arg\max}\limits_{y_0\in O;x_0,x_1\in E}\left[\delta_{0,1}(y_0,x_0,x_1)p_{x_{1}\rightarrow x_2}\right]$\\
  \tcc{Now propagate maximums, keeping track where they occur.}
  \For {n:=3 to N}{
$\delta_n(x_n)=\max\limits_{x_{n-1}\in E}\left[\delta_{n-1}(x_{n-1})p_{x_{n-1}\rightarrow x_n}\right]q_{Y_{n-1}\rightarrow Y_n}(x_n)$
\tcp*{Maximums}
  $\psi_n(x_n)=\mathop{\arg\max}\limits_{x_{n-1}\in E}\left[\delta_{n-1}(x_{n-1})p_{x_{n-1}\rightarrow x_n}\right]$
\tcp*{Maximum Locations}
}  
   \tcc{Termination}
  $P^*=\max\limits_{x_{N}\in E}\left[\delta_{N}(x_{N})\right]$\\
  $x^*_N=\mathop{\arg\max}\limits_{x_{N}\in E}\left[\delta_{N}(x_{N})\right]$\\
    \tcc{Path Back Tracking}
  \For {n:=N-1 down to 2}{$ x^*_n=\psi_{n+1}(x^*_{n+1})$}
$(y^*_0;x^*_0,x^*_1)=\psi_{2}(x^*_{2})$
\caption{Viterbi algorithm for MOM}
\end{algorithm}
\subsection{Dynamic Programming Explanation}
$\delta_2(x)$ as defined in Algorithm
\ref{alg2}  verifiably satisfies (\ref{deltaver})
by simple substitution.
Next, assume $\delta_{n-1}(x)$ satisfies (\ref{deltaver}) for some $n\ge 3$.  
Then, by the algorithm and the Markov property:
\begin{eqnarray} 
&&\ \ \ \ \ \delta_n(x_n)\\\nonumber
&\!\!\!=&\!\!\!\max\limits_{x_{n-1}\in E}\!\left[\delta_{n-1}(x_{n-1})p_{x_{n-1}\rightarrow x_n}\right]q_{Y_{n-1}\rightarrow Y_n}(x_n)\\\nonumber
&\!\!\!=&\!\!\!\max\limits_{x_{n-1}}\!\left[\max_{y_0;x_0,...,x_{n-2}}\!\!P(Y_0=y_0;X_0=x_0,...,X_{n-1}=x_{n-1};Y_1,...,Y_{n-1})p_{x_{n-1}\rightarrow x_n}\!\right]
\!q_{Y_{n-1}\rightarrow Y_n}(x_n)\!\!\!\!\!\!\!\!\! \!\!  \\\nonumber
&\!\!\!=&\!\!\!\max_{y_0;x_0,...,x_{n-2},x_{n-1}}\!P(Y_0=y_0;X_0=x_0,...,X_{n-2}=x_{n-2},X_{n-1}=x_{n-1};
X_n=x_n;Y_1,...,Y_{n}),
\end{eqnarray} 
and (\ref{deltaver}) follows for all $n$ by induction.
Next, it follows from the algorithm that $x^*_N=\mathop{\arg\max}\limits_x\delta_N(x)$.
Finally, we have by the Path Back Tracking part of the algorithm as well as induction that
\begin{eqnarray} 
\delta_N(x^*_N)&\!=&\!\delta_{N-1}(x^*_{N-1})p_{x^*_{N-1}\rightarrow x^*_N}q_{Y_{N-1}\rightarrow Y_N}(x^*_N)\\\nonumber
&\!=&\!\delta_{0,1}(y_0^*;x^*_0,x^*_{1})\prod_{n=2}^N p_{x^*_{n-1}\rightarrow x^*_n}q_{Y_{n-1}\rightarrow Y_n}(x^*_n)\\\nonumber
&\!=&\!P(Y_0=y_0^*;X_0=x^*_0,X_1=x^*_1,...,X_{N-1}=x^*_{N-1};
X_N=x^*_N;Y_1,...,Y_{N})
\end{eqnarray} 
and the most likely sequence is established.

\begin{remark} 
Our Viterbi dynamic programming algorithm can be thought of as 
a direct generalization of the orginal Viterbi algorithm for HMM.
Indeed, we need only let $q_{y\rightarrow y'(x)}= b_x(y')$ for some
probability mass function (depending upon $x$) $b_x$ to recover the 
normal HMM and the normal Viterbi algorithm.
Then, we would drop the consideration of the most likely starting
point $(y_0^*,x_0^*)$ and be back to the original setting.
\end{remark}

\subsection{Small Number Problem}

The Viterbi-type algorithm also suffers from extraordinarily small,
shrinking numbers.
Indeed, since we have multiple events in both $X$ and $Y$, the
numbers will shrink faster than our EM algorithm in $n$.
On the other hand, we are not taking ratios and it is easier to
scale this algorithm than the EM algorithm.
Still, one might wonder if we can handle the small number problem
for our dynamic programming algorithm in a similar manner as we
did for our EM algorithm.

While we did not adjust Algorithm \ref{alg2}, this algorithm
can be adjusted for small numbers.
The idea is similar to that used in the EM algorithm.
Simply replace $\delta_2$ with
$$
\upsilon_2(x_2)=\max\limits_{y_0\in O;x_0,x_1\in E}\left[\delta_{0,1}(y_0,x_0,x_1)p_{x_{1}\rightarrow x_2}\right]q_{Y_{1}\rightarrow Y_2}(x_2)
$$
$$
\gamma_2(x_2)=\frac{\upsilon_2(x_2)}{a_2},\  a_2=\sum\limits_{\xi}\upsilon_2(\xi)
$$
and $\delta_n,\ n\ge 3$ with $\gamma_n$, where
$$
\upsilon_n(x_n)=\max\limits_{x_{n-1}\in E}\left[\gamma_{n-1}(x_{n-1})p_{x_{n-1}\rightarrow x_n}\right]q_{Y_{n-1}\rightarrow Y_n}(x_n)
$$
$$
\gamma_n(x_n)=\frac{\upsilon(x_n)}{a_n},\  a_n=\sum\limits_{\xi}\upsilon_n(\xi).
$$
Then, replace $\delta_n$ with $\gamma_n$ everywhere else in the algorithm.
Of course, the maximum sequence likelihood $P^*$ would have be scaled
down by multiplying by the product of the $a_n$'s.
Otherwise, the algorithm would remain the same.

\section{Bitcoin Example}\label{BitCoin}

To establish the applicability of our model and algorithms to real-world
big data problems,
we include an illustrative MOM model example application.
In particular, no Gaussian approximation is imposed.
We work with discrete data and rely on our solution to the small
number problem.
Bitcoin is a highly volatile digital currency that can be
traded by various means.
Further, holding Bitcoin during uptrends has proven to be a superlative
investment, while holding it during other periods has been extremely
risky and painful.
Therefore, it is of interest to see if our MOM model algorithms might be able
to isolate uptrend periods and provide a two-hidden-state Markov Observation
Model that matches historical data reasonably well.
Accordingly, we applied our (Baum-Welch-like) EM and 
(Viterbi-like) dynamic programming algorithms to
identify and demonstrate a (hidden) regime-change model for
daily Bitcoin closing prices from Sept 1, 2018 until Sept 1, 2022.

\begin{figure}
    \centering
    \includegraphics{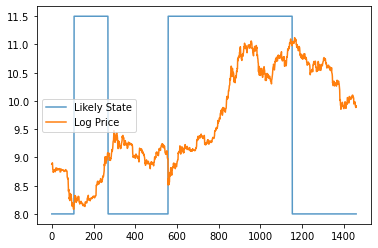}
\begin{tabular}{r@{: }l r@{: }l}
Blue High & Most likely uptrend; &
Blue Low & Anything but an uptrend. 
\end{tabular}
\caption{BitCoin Uptrend Detection - Three Uptrends}
\label{BitcoinFig}
\end{figure}

The price varied (rather dramatically) from a low of $3235.76$ USD
on Dec. 15, 2018 to a high of $67566.83$
USD on Nov. 8, 2021 over our four year period of interest.
Instead of raw prices, we took our observations $Y$ be the natural logarithm 
of prices, which ranged from $8.0823$ to $11.12$ (see the continuous
orange line in Figure \ref{BitcoinFig}), 
and divided those into $b=25$ equal-sized bins.
For example, bin $0$ consisted of log prices with the range
$8.08$ to $8.2016$ corresponding to actual prices \$$3230$ to \$$3646.79$
while the last bin, bin $24$, consisted of log prices with the range $11.0005$ to
$11.122$ corresponding to actual prices \$$59904.09$ to \$$67643.06$ all in 
US dollars. 
For ease of assimilation we just chose 
to have $s=2$ hidden states $0$ and $1$.

\subsection{Initialization}
It is well known that the Baum-Welch algorithm for HMM will get
stuck at a local maximum.
This should be even more true for our EM algorithm of our MOM model
as we have even more to estimate.
(MOM has a larger initial distribution and more complex Markov transitions
probabilities compared to HMM's single state initial distribution and 
emission probabilities.)
Therefore, it makes sense to start the algorithm with an idea of
the solution that we seek.
Most importantly, we want to differentiate the hidden states so
we plan that state $1$ will represent an uptrend and state $0$ will
represent everything else and initialize accordingly.
However, since we want our algorithm to find a variety of uptrends,
we will allow some inconsistencies in our initial set up that will force
the algorithm to make significant changes.
Also, we recognize that it is the algorithm that decides what the
hidden states are.
While we suggesting state $1$ will be an uptrend, the algorithm, by
the time it has finished, may have decided $1$ represents something 
completely different like high volatility say.

Our first step was to use the data to come up with initial 
$q_{y\rightarrow y'}(1)$ for uptrend observation transitions 
and $q_{y\rightarrow y'}(0)$ other observation transitions.
Accordingly, we made a somewhat arbtrarily decision about when Bitcoin
might be in an uptrend.
In particular, we decided, based on a brief glimpse at the graph,
to say it was in an uptrend from the low on 
December 15, 2018 until the high on July 3, 2019, then again
from the low on March 12, 2020 until the high on April 15, 2021,
and finally from the low on July 20, 2021 until the high on November 8, 2021.
Otherwise, it was not in an uptrend.
This amounts to six changes over the $1461$ days
in these four years.

\begin{remark}
Naturally, there were down and up days for both hidden states.  
Also, the very first price was excluded as this is our $Y_0$ price that
we would not see in practice.
Finally, we need to emphasize that we expect that the bin size effect 
was rather huge.
We used $25$ bins, which is extremely crude, and an arbitrary $4$ year
period with no sign of numerical issues.
Also, the amount of data was very uneven over the bins, which we simply
ignored, but it certainly hampered algorithm performance.
A larger, finer study by more experienced computer programmers is definitely
recommended.
\end{remark}

To create our initial observation transitions, we initalized our $b \times b$ matrices 
$\{q_{y\rightarrow y'}(0)\}$ and $\{q_{y\rightarrow y'}(1)\}$
to zero.
(Here, $y$ and $y'$ refer to either bin number or rounded log price through a
one-to-one mapping.)
Starting from $n=1$ and going through to $n=N-1$ we added $1$ to
$q_{Y_n\rightarrow Y_{n+1}}(1)$
if we were in an uptrend and otherwise $1$ to $q_{Y_n\rightarrow Y_{n+1}}(0)$.
Then, we normalized both matrices so that the non-zero rows added to one.

To initialize $\mu(x,y)$, we first set it all to zero.
Next, we went through $n\in\{1,...,N-1\}$
if $Y_{n+1}$ was equal to $Y_1$ then we added $q_{Y_n\rightarrow Y_1}(x)$
to $\mu(x,Y_n)$ for $x=0,1$.
Finally, we normalized $\mu(x,y)$ so it summed to $1$.

We set the stopping criterion to be extremely tight, making sure that
the $p$'s and $\mu$'s were essentially done changing.
(The $q$'s will also be done in this case so there is little need
to check this bulky matrix.)
The initialization of the $p$ will be varied and explained in the results.

\subsection{Results}

Our first goal was to see if the algorithms would return the
three uptrends that were supplied.
To do this, we initialized $p$ as follows:
\begin{equation}\displaystyle
\left[\begin{array}{cc}p_{0\rightarrow 0}&p_{0\rightarrow 1}\\p_{1\rightarrow 0}&p_{1\rightarrow 1}\end{array}\right]=
\left[\begin{array}{cc}0.997&0.003\\0.003&0.997\end{array}\right].
\end{equation}  
This means that it should switch states every $333$ days on average, 
which is roughly consistent with my initial take of three uptrends
given above.

After $k=11$ iterations, the EM algorithm converged and the
combined result of both algorithms is displayed in Figure \ref{BitcoinFig}.
It reduced the number of uptrends from what I supplied from three to two.
I believe that the algorithm's uptrends are at least as good as my initial
ones.
The final $p$ matrix in this case was
\begin{equation}\displaystyle
\left[\begin{array}{cc}p_{0\rightarrow 0}&p_{0\rightarrow 1}\\p_{1\rightarrow 0}&p_{1\rightarrow 1}\end{array}\right]=
\left[\begin{array}{cc}0.99643132&0.00356868\\0.00302665&0.99697335\end{array}\right].
\end{equation}  
\begin{figure}
    \centering
    \includegraphics{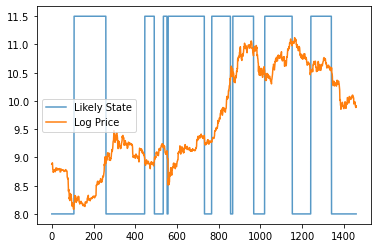}
\begin{tabular}{r@{: }l r@{: }l}
Blue High & Most likely uptrend; &
Blue Low & Anything but an uptrend. 
\end{tabular}
\caption{BitCoin Uptrend Detection - Short Uptrends}
\label{BitcoinFiga}
\end{figure}

From an investor's perspective shorter, steeper uptrends might be more
desirable.
Hence, we investigated the possibility of finding more, shorter uptrends
without retraining the $q$ matrices.
Instead, we merely changed the initial $p$ matrix to\begin{equation}\displaystyle
\left[\begin{array}{cc}p_{0\rightarrow 0}&p_{0\rightarrow 1}\\p_{1\rightarrow 0}&p_{1\rightarrow 1}\end{array}\right]=
\left[\begin{array}{cc}0.90&0.10\\0.01&0.99\end{array}\right],
\end{equation}  
which initially makes all changes more likely.
However, it sets the initial expected time in an uptrend to just ten days initially. 
After $k=43$ iterations, the EM algorithm converged and the
combined result of both algorithms is displayed in Figure \ref{BitcoinFiga}.
Compared to the earlier result the uptrends were split and shrunk.
In addition, a new uptrend was added in the later part of the data stream.
It is very interesting that it did a decent job of finding a different type
of uptrend without any new training, but rather just a different
$p$ matrix initialization.
The final $p$ matrix was:
\begin{equation}\displaystyle
\left[\begin{array}{cc}p_{0\rightarrow 0}&p_{0\rightarrow 1}\\p_{1\rightarrow 0}&p_{1\rightarrow 1}\end{array}\right]=
\left[\begin{array}{cc}0.98329893&0.01670107\\0.0127778&0.9872222\end{array}\right].
\end{equation} 
The EM algorithm spent those $43$ iterations making signficant changes.
In particular, the final $p$ matrix and graph suggests a near equal time
in uptrends as not.
However, this is somewhat out of our control.
We supply the data, the number of hidden states and some initial
estimates and then tell the EM algorithm to give us the 
locally optimal model, whatever that may be.
In both cases the result exceeded our expectations.

\bibliographystyle{apalike}

\bibliography{DMOM}
\bigskip

\end{document}